%% file: thesis.tex
\documentclass[a4paper,12pt,times,numbered,print,oneside,index]{Classes/PhDThesisPSnPDF}

\usepackage[figuresright]{rotating}
\input{Preamble/preamble}

\input{thesis-info}

\ifdefineAbstract
\fi

\ifdefineChapter
\fi

\begin{document}

\frontmatter

\maketitle

\include{Dedication/dedication}
\include{Acknowledgement/acknowledgement}
\pagenumbering{roman}
\include{Abstract/abstract}


\tableofcontents
\listoffigures
\listoftables

\printnomenclature

\mainmatter

\begin{spacing}{1.5}
\captionsetup[subfigure]{labelformat=empty}

\include{Chapter1/chapter1}
\include{Chapter2/chapter2}
\include{Chapter3/chapter3}
\include{Chapter4/chapter4}
\include{Chapter5/chapter5}

\end{spacing}

\bibliographystyle{unsrt} 
\cleardoublepage
\bibliography{References/references} 









\begin{appendices} 

\end{appendices}

\printthesisindex 

\end{document}

%% file: thesis-info.tex
\title{Local Area Transform for Cross-Modality Correspondence Matching and Deep Scene Recognition}

\author{Seungchul Ryu}

\dept{Department of Electrical and Electronic Engineering}

\university{Yonsei University}

\degreetitle{Doctor of Philosophy}

\degreedate{February 2019} 

\subject{LaTeX} \keywords{{LaTeX} {PhD Thesis} {Engineering} {Yonsei University}}

%% file: Dedication/dedication.tex

\begin{dedication} 
\vskip 2cm
I would like to dedicate this thesis to my loving family \dots

\end{dedication}

%% file: Acknowledgement/acknowledgement.tex

\begin{acknowledgements}      
I would like to express my sincere gratitude to my supervisor Prof. Kwanghoon Sohn for the continuous support of my Ph.D sutdy and related research, for his patience, motivation, and immense knowledge. Hist guidance helped me in all the time of research and writing of this dissertation. Besides my supervisor, I would like to thank my dissertation committe: Prof. Euntae Kim, Prof. Hyeran Byun, Prof. Sangyoun Lee, and Prof. Dongbo Min, for their insightful comments and encouragement, but also for the hard question which incented me to widen my research from various perspectives. My sincere thanks also goes to Dr. Jungdong Seo, Dr. Donghyun Kim, Prof. Bumsub Ham, who provided me an insights about my research. Without their precious support it would not be possible to conduct this research. I thank my fellow lab-mates, Dr. Seungryong Kim, Dr. Changae Oh, Dr. Youngjung Kim, Kihong Park, and Sunok Kim in for the discussions and for all the fun we have had in the last years. Also, I thank Dr. Cho who provided me an opportunity to join their great team. Last but not the least, I would like to thank my family: my parents, my brother, my wife, and my daughters for supporting me spiritually throughout writing this dissertation and my life in general.\\

\end{acknowledgements}

%% file: Abstract/abstract.tex
\begin{abstract}
Establishing correspondences is a fundamental task in variety of image processing and computer vision applications. In particular, finding the correspondences between a non-linearly deformed image pair induced by different modality conditions is a challenging problem. This paper describes a efficient but powerful image transform called local area transform (LAT) for modality-robust correspondence estimation. Specifically, LAT transforms an
image from the intensity domain to the local area domain, which is invariant under nonlinear intensity deformations, especially radiometric, photometric, and spectral deformations. In addition, robust feature descriptors are reformulated with LAT for several practical applications. Furthermore, LAT-convolution layer and Aception block are proposed and, with these novel components, deep neural network called LAT-Net is proposed especially for scene recognition task. Experimental results show that LATransformed images provide a consistency for nonlinearly deformed images, even under random intensity deformations. LAT reduces the mean absolute difference by approximately 0.20 and the different pixel ratio by approximately 58\% on average, as compared to conventional methods. Furthermore, the reformulation of descriptors with LAT shows superiority to conventional methods, which is a promising result for the tasks of cross-spectral and modality correspondence matching. LAT gains an approximately 23\% improvement in the
correct detection ratio and a 10\% improvement in the recognition rate for the tasks of RGB-NIR cross-spectral template matching and cross-spectral feature matching, respectively. LAT reduces the bad pixel percentage by approximately 15\% and the root mean squared errors by 13.5 in the task of cross-radiation stereo matching. LAT also improves the cross-modal dense flow estimation task in terms of warping error, providing 50\% error reduction. LAT-Net provides 14\% and 7\% accuracy improvements in cross spectral scene recognition and domain generalized scene recognition tasks, respectively. the local area can be considered as an alternative domain to the intensity domain to achieve robust correspondence matching, image recognition, and a lot of applications: such as feature matching, stereo matching, dense correspondence matching, image recognition, and image retrieval.
\end{abstract}

%% file: Chapter1/chapter1.tex

\chapter{Introduction}  

Correspondence matching is a basic and fundamental task in a vast range of image processing and computer vision applications: image denoising \cite{app1, trinh2014novel}, image editing \cite{app2, bugeau2014variational}, object tracking \cite{app3}, stereo matching \cite{app4}, optical flow \cite{revaud2015epicflow}, image retrieval \cite{babenko2015aggregating}, image recognition \cite{deng2009imagenet, uijlings2013selective}, and scene recognition \cite{kwitt2012scene, su2012improving}. Conventional correspondence matching algorithms are commonly based on gradient-based descriptors \cite{SIFT, bay2008speeded, HOG}. In real world, however, images are acquired in an uncontrolled environment; thus, the image may suffer intensity deformations due to changes in illumination conditions, camera photometric parameters, viewing positions, and so on \cite{problem}. Furthermore, recently, cross-modality imaging system (e.g., multi-spectral imaging system \cite{DB1, sorensen2015multimodal} has been attracted many attentions to address challenging problems occurring in the conventional unimodal imaging system. Images acquired from different modalities also have intensity deformations due to changes in sensor responses and spectral distributions.

These deformations between patches or images induce the inaccuracy problem of the correspondence matching. Let ${{\bf{I}}_1}$ and ${{\bf{I}}_2}$ be two input images, and $\alpha({\bf{p}}) \in {\bf{I}}_2$ be the corresponding pixel of ${\bf{p}} \in {\bf{I}}_1$. When dealing with a correspondence matching under uncontrolled environments or multi-modalities, three groups of approaches have been considered: tone mapping, color constancy, and robust similarity measure. The first group, called tone mapping, attempts to determine a mapping function $\mathcal M$ such that $\mathcal M \{ {{\bf{I}}_1}({\bf{p}})\}  = {{\bf{I}}_2}({\alpha ({\bf{p}})})$. A classic method for extracting $\mathcal M$ is a histogram matching \cite{HistogramMatching}, which computes a mapping function that optimally aligns the histogram of ${\bf{I}}_1$ with that of ${\bf{I}}_2$. Several methods compute a mapping function $\mathcal M$ based on the statistical distribution of intensity values \cite{Statistical1}. More sophisticated mapping functions were well reviewed in \cite{ToneMappingReview2}. Tone mapping approaches commonly assume that ${\bf{I}}_1$ and ${\bf{I}}_2$ are entirely aligned into same scene regions. This assumption is clearly hold only when the images are taken at the same viewpoint under the same illumination condition, but in other cases the obtained mapping function $\mathcal M$ might be erroneous and inconsistent.

The second group, called color constancy, tries to find a model $\mathcal S$ to transform images into constant color space removing illumination components such that $\mathcal S \{ {{\bf{I}}_1}({\bf{p}})\}  = \mathcal S \{{{\bf{I}}_2}({\alpha ({\bf{p}})})\}$. One of the most popular methods is grey-world model which removes the illumination spectral distribution factor with an assumption that, under a white light source, the average color in a scene is achromatic (i.e., grey) \cite{NormalizedChromaticity}. Another well-known method, white patch retinex model, assumes that the maximum response in an image is caused by a perfect reflectance (i.e., white patch). In practice, this assumption is alleviated by considring the color channels separately, resulting in the max-RGB algorithm. The normalized chromaticity model is commonly used for the elimination of the lighting geometry factors under the Lambertian reflectance model \cite{NormalizedChromaticity}. Gamut mapping and other learning based algorithms have been also investigated \cite{LearningColorConstancy2}. However, most models cannot remove the dependency of the lighting geometry and the illumination spectral distribution simultaneously as will be discussed in Chapter \ref{chap2}.

The third group, called robust similarity measure, attempts to describe a local signature within a patch invariant to a nonlinear deformation. In some cases, an intensity deformation is nonlinear but still maintains a monotonicity, i.e., the order of intensity-levels is preserved. Similarity measures based on such an ordinal value include local binary pattern (LBP) \cite{LBP}, binary robust independent elementary features (BRIEF) \cite{BRIEF}, rank transform (RT) \cite{Rank}, and census transform (CT) \cite{census}. Although these ordinal information based approaches account for a monotonic mapping, they fail under a non-monotonic intensity deformation.

A gradient-based similarity measure, such as histogram of gradients (HOG) \cite{HOG} and scale invariant feature transform (SIFT) \cite{SIFT} has been considered a photometric invariant similarity measure. Such a method inherently, however, causes the loss of information due to the contraction of data weakening their discrimination power and fails under a non-monotonic mapping. Normalized cross correlation (NCC) measures the cosine of an angle between two vectors, and thus is robust to a linear intensity deformation. To address the inaccuracy at an object boundary of the NCC, adaptive normalized cross correlation (ANCC) is proposed in \cite{ANCC}. In \cite{MTM}, a generalized version of NCC is proposed, which is called matching by tone mapping (MTM). Mutual information (MI) \cite{MI} is widely used similarity measure for images with nonlinear deformations. MI measures the statistical dependence between two vectors $v_1$ and $v_2$ by computing the loss of entropy in $v_1$ given $v_2$.

To summarize, the conventional methods approached to solve the problem of a nonlinear intensity deformation by adjusting intensity values to be similar or utilizing a gradient, ordinal information, and a statistical measure. However, these approaches cannot account for a general nonlinear intensity deformation. This paper proposes to use \emph{local area information} as a robust index for nonlinear intensity deformations. We define local area transform (LAT) as a robust mapping of an image from an intensity domain to a local area domain. LAT is designed to address the nonlinear deformation problem of images which may be acquired from different photometric parameters, light sources, and modalities. The objective of LAT is similar to a color constancy, i.e., transferring an image from the original intensity (or color) values to constant intensity (or color) domain. However, unlike the color constancy LAT alters an image from intensity domain, which is sensitive to a nonlinear deformation, to robust local area domain. Ordinal transform such as LBP, RT, and CT also aims to transfer an image to ordinal information domain, but fails under a non-monotonic intensity deformation. As our knowledge, this study is the first attempt to address a nonlinear deformation problem with the local area information in the task of a correspondence matching.

This study prove that the LAT is robust image transform for non-linear intensity, radiometric, photometric, and spectral deformations. Also, efficient implementation of LAT is proposed with integral histogram. Besides the use as a transformation, the concept of LAT is extended to reformulate the conventional robust feature descriptors such as SIFT, LSS, CT, RT, and etc. The reformulation embeds great properties of LAT into the conventional feature descriptors. The reformulated descriptors show that superior performance in tasks of non-linear deformation correspondence matching, cross-spectral correspondence matching, cross-radiometry stereo matching, and cross-modality dense correspondence matching. Furthermore, novel deep networks are proposed to address cross-domain scene recognition problem. In the proposed deep scene recognition network, conventional convolutional layers are replaced by LAT-convolution layers and aception block is introduced. The proposed deep scene recognition networks outperform the conventional methods in tasks of cross-spectral scene recognition and domain generalized scene recognition.

The remainder of this dissertation is organized as follows. In Chapter \ref{chap2}, related literatures are presented. In Chapter \ref{chap3}, LAT is described with its properties and implementation details. LAT-reformulated features and LAT-Net are also presented. In Chapter \ref{chap4}, the performances of LAT are evaluated in tasks of nonlinear-deformed image matching, cross spectral correspondence matching, cross radiometry stereo matching, cross modal dense flow estimation, and cross modality scene recognition. Chapter \ref{chap5} concludes this paper with the discussions.

%% file: Chapter2/chapter2.tex
\chapter{Related Works}\label{chap2}  
An image taken by a linear imaging device with $i^{th}$ sensor is modeled as \cite{ImageModel}:
\begin{equation}\label{eq1}
{{\bf{I}}^i}({\bf{p}}) = \int_\omega  {E(T,\lambda )S({\bf{p}},\lambda)F_i(\lambda )d\lambda },
\end{equation}
where ${{\bf{I}}^i}({\bf{p}})$ denotes the sensor response at a point $\bf{p}$ in the spatial coordinate, $E(T,\lambda)$ represents the spectral distribution of the incident illuminant, $S({\bf{p}},\lambda)$ represents the surface reflectance at $\bf{p}$, and $F_i(\lambda)$ represents the spectral response of the sensor. Approximating the sensor spectral response $F_i(\lambda)$ as the Dirac delta function such that $F_i(\lambda)=\upsilon _i \delta (\lambda-\lambda_i)$, (\ref{eq1}) is simplified as follows:
\begin{equation}\label{eq2}
{{\bf{I}}^i}({\bf{p}}) = E({T, \lambda_i})S({\bf{p}},{\lambda_i}){\upsilon_i}.
\end{equation}

Under Planck's law, the spectral distribution of the illuminant $E(T,\lambda_i)$ is modeled a function of the absolute temperature $T$ and the wavelength $\lambda$ as $E(T,\lambda_i) = {c_1}{{\lambda_i}^{ - 5}}{e^{{{{c_2}} \mathord{\left/ {\vphantom {{{c_2}} {\lambda_i T}}} \right. \kern-\nulldelimiterspace} {\lambda T}}}}$ where ${c_1} \buildrel \Delta \over = 2h{c^2}$, ${c_2} \buildrel \Delta \over = \frac{{hc}}{k}$, $c$ is the speed of light, $h$ is Planck's constant, and $k$ is Boltzmann constant. The surface reflectance $S({\bf{p}},\lambda_i)$ is represented as $S({\bf{p}},\lambda_i)=m({\bf{p}})S_m({\bf{p}},\lambda_i)$ where $m(\bf{p})$ is a lighting geometry factor and $S_m({\bf{p}},\lambda_i)$ is the matte-surface reflectance with the assumption of a matte surface. Taking the exposure time $\varepsilon_i$ into the consideration, the image acquisition model in (\ref{eq2}) is modified as
\begin{equation}\label{eq3}
{{\bf{I}}^i}({\bf{p}}) = {\varepsilon_i}E({T, \lambda_i})S({\bf{p}},{\lambda_i}){\upsilon_i}.
\end{equation}

When images are acquired in an uncontrolled environment or in cross-modality system, they suffer from nonlinear deformation problem induced by different modalities. To address the correspondence problem under uncontrolled environments or multi-modalities, three groups of approaches have been explored: tone mapping, color constancy, and robust similarity measure. Color constancy is closely related works to the proposed LAT. Color constancy tries to find a model $\mathcal S$ to transform images into constant color space removing illumination components such that $\mathcal S \{ {{\bf{I}}_1}({\bf{p}})\}  = \mathcal S \{{{\bf{I}}_2}({\alpha ({\bf{p}})})\}$. One of the most popular methods is grey-world model which removes the illumination spectral distribution factor with an assumption that, under a white light source, the average color in a scene is achromatic (i.e., grey) \cite{NormalizedChromaticity}. Another well-known method, white patch retinex model, assumes that the maximum response in an image is caused by a perfect reflectance (i.e., white patch). In practice, this assumption is alleviated by considring the color channels separately, resulting in the max-RGB algorithm. The normalized chromaticity model is commonly used for the elimination of the lighting geometry factors under the Lambertian reflectance model \cite{NormalizedChromaticity}. Gamut mapping and other learning based algorithms have been also investigated \cite{LearningColorConstancy2}. However, most models cannot remove the dependency of the lighting geometry and the illumination spectral distribution simultaneously. More recently, deep neural networks based color constancy methods were also explored \cite{bianco2015color, barron2015convolutional, oh2017approaching, hu2017fc4}

Grey world model estimates the illuminant by averaging channel values under the assumption that the average reflectance in an image is achromatic, and is proven to be an instantiation of Minkowski-norm ($\rho=1$) \cite{Minkowski}. Then, the gray world model ${{\bf{I}}_G}^i({\bf{p}})$ is computed as follows:
\begin{equation}\label{eq4}
\begin{array}{l}
{{\bf{I}}_G}^i({\bf{p}}) = {{\bf{I}}^i}({\bf{p}})/\sum\limits_{{\bf{q}}} {{{\bf{I}}^i}({\bf{q}})} \\
\quad \quad \;\; = {\varepsilon_i}E(T,{\lambda _i})S({\bf{p}},{\lambda _i}){\upsilon _i}/\sum\limits_{{\bf{q}}} {{\varepsilon_i}E(T,{\lambda _i}){\upsilon _i}S({\bf{q}},{\lambda _i})}.
\end{array}
\end{equation}

In practice, it is computed within local neighbors $\mathcal N_{\bf{p}}$ with the assumption of $E(T,\lambda_i)$ to be locally constant, thus (\ref{eq4}) is simplified as:
\begin{equation}\label{eq5}
{{\bf{I}}_G}^i({\bf{p}}) = S({\bf{p}},{\lambda _i})/\sum\limits_{{\bf{q}} \in \mathcal N_{\bf{p}}} {S({\bf{q}},{\lambda _i})}.
\end{equation}

(\ref{eq5}) implies that the gray world model is invariant to an illumination deformation under the local-constancy assumption. However, when dealing with images acquired by different modalities (e.g., cross-spectral) $S$ undergoes non-linear deformation, thus the gray world model is no longer guarantee the robustness to spectral deformations.

The $k^{th}$-channel normalized chromaticity ${{\bf{I}}_N}^k({\bf{p}})$ \cite{NormalizedChromaticity} eliminates the effect of the lighting geometry by dividing each channel response by the average of them as follows:
\begin{equation}
{{\bf{I}}_N}^k({\bf{p}}) = {{\bf{I}}^k}({\bf{p}})/\sum\limits_{j \in (1,n)} {{{\bf{I}}^j}({\bf{p}})},
\end{equation}
where $n$ is the number of channels. Substituting (3), (6) is simplified as
\begin{equation}
{{\bf{I}}_N}^k({\bf{p}}) = \frac{{\varepsilon_k E(T,{\lambda _k}){S_m}({\bf{p}},{\lambda _k}){\upsilon _k}}}{{K({\bf{p}})}},
\end{equation}
where $K({\bf{p}}) = \sum\limits_{j \in (1,n)} {\varepsilon_k E(T,{\lambda _j}){S_m}({\bf{p}},{\lambda _j}){\upsilon _j}}$. (7) indicates that the normalized chromaticity only removes the lightning geometry factor $m(\bf{p})$. Log-chromaticity \cite{ANCC} defined as ${{\bf{I}}_l}^k({\bf{p}}) = \log ({{\bf{I}}^k}({\bf{p}})/\sqrt[n]{{\prod\limits_{j \in (1,n)} {{{\bf{I}}^j}({\bf{p}})} }})$ transforms a nonlinear deformation into a linear deformation. However, both the normalized chromaticity and the log-chromaticity cannot be applicable to uni-channel image, e.g., infra-red image.

Tone mapping algorithms attempt to construct a mapping function $\mathcal M$ such that $\mathcal M \{ {{\bf{I}}_1}({\bf{p}})\}  = {{\bf{I}}_2}({\alpha ({\bf{p}})})$. A classic method for extracting $\mathcal M$ is a histogram matching \cite{HistogramMatching}, which computes a mapping function that optimally aligns the histogram of ${\bf{I}}_1$ with that of ${\bf{I}}_2$. Several methods compute a mapping function $\mathcal M$ based on the statistical distribution of intensity values \cite{Statistical1, eilertsen2015real}. More sophisticated mapping functions were well reviewed in \cite{ToneMappingReview2}. Tone mapping approaches commonly assume that ${\bf{I}}_1$ and ${\bf{I}}_2$ are entirely aligned into same scene regions. This assumption is clearly hold only when the images are taken at the same viewpoint under the same illumination condition, but in other cases the obtained mapping function $\mathcal M$ might be erroneous and inconsistent. Histogram matching, the most common tone mapping scheme, aligns the histogram of ${\bf{I}}_1$ to that of ${\bf{I}}_2$ when they are acquired from the same scene at the same viewpoint, i.e., $\alpha({\bf{p}})={\bf{p}}$. However, this assumption is too hard to be applied to practical environments. In addition, the histogram matching is stable only for global deformations, and is no longer guarantees for local deformations.

Robust similarity measure attempts to describe a local signature within a patch invariant to a nonlinear deformation. In some cases, an intensity deformation is nonlinear but still maintains a monotonicity, i.e., the order of intensity-levels is preserved. Similarity measures based on such an ordinal value include local binary pattern (LBP) \cite{LBP}, binary robust independent elementary features (BRIEF) \cite{BRIEF}, rank transform (RT) \cite{Rank}, and census transform (CT) \cite{census}. Although these ordinal information based approaches account for a monotonic mapping, they fail under a non-monotonic intensity deformation.

A gradient-based similarity measure, such as histogram of gradients (HOG) \cite{HOG} and scale invariant feature transform (SIFT) \cite{SIFT} has been considered a photometric invariant similarity measure. Such a method inherently, however, causes the loss of information due to the contraction of data weakening their discrimination power and fails under a non-monotonic mapping. Recently, dense adaptive self-correlation (DASC) descriptor has been proposed to provide robustness for modality variations, but is also has limitations on non-linear deformations \cite{kim2015dasc}. Normalized cross correlation (NCC) measures the cosine of an angle between two vectors, and thus is robust to a linear intensity deformation. To address the inaccuracy at an object boundary of the NCC, adaptive normalized cross correlation (ANCC) is proposed in \cite{ANCC}. In \cite{MTM}, a generalized version of NCC is proposed, which is called matching by tone mapping (MTM).  Mahalanobis distance cross-correlation (MDCC) has also been proposed \cite{kim2014mahalanobis}. Mutual information (MI) \cite{MI} is widely used similarity measure for images with nonlinear deformations. MI measures the statistical dependence between two vectors $v_1$ and $v_2$ by computing the loss of entropy in $v_1$ given $v_2$. Recently, deep learning based similarity measure is also actively studied \cite{chen2015deep, kim2017fcss, han2017scnet, ufer2017deep}

Under a linear deformation written as ${{\bf{I}}_2}(\alpha ({\bf{p}})) = a\,{{\bf{I}}_1}({\bf{p}}) + a'$ where $a$ and $a'$ are constants, a gradient is deformed with a scaling factor $a$: $\Delta {{\bf{I}}_2}(\alpha ({\bf{p}})) = a\,\Delta {{\bf{I}}_2}({\bf{p}})$, thus gradient information can be a robust feature when $a>0$. However, when $a<0$ the gradient inversion occurs, which leads the inaccuracy of gradient based similarity measures such as HOG and SIFT. When the deformation is non-linear, the gradients fail to be preserved across the deformation. In some cases, the intensity deformation is nonlinear but still maintains monotonicity, i.e., the order of intensity-levels is preserved as $\forall {\bf{p}},{\bf{q}}\;\;{\rm{if}}\;{{\bf{I}}_1}({\bf{p}}) \le {{\bf{I}}_1}({\bf{q}}),\;\;{{\bf{I}}_2}(\alpha ({\bf{p}})) \le {{\bf{I}}_2}(\alpha ({\bf{q}}))$. An intensity ordinal similarity measure, such as LBP, RT, and CT, provides the robustness under the assumption of the monotonicity, but the assumption is violated in a general non-linear deformation. The local intensity order is not preserved across non-linear deformation, thus which leads the inaccuracy of an intensity ordinal similarity measure under the non-linear deformation.

One of the most important application in computer vision is image recognition. Especially, scene image recognition is an important problems for applications of computer vision such as robotics, image search, geo-localization, etc. However, scene recognition is challenging problem because scenes commonly include both a holistic component and object-based components. Conventional methods for scene recognition can be categorized into holistic gist descriptors \cite{oliva2001modeling} and local feature based descriptors \cite{nowak2006sampling}. Local feature based approaches were mainly based on bag-of-features (BoF) representation, using local features such as SIFT or HOG \cite{kwitt2012scene, li2010object, su2012improving}, combined through a pooling operator. Sophisticated pooling strategies such as the vector of locally aggregated descriptors (VLAD) \cite{su2012improving} or the Fisher vector (FV) \cite{sanchez2013image} emerged as the dominant mechanism for scene recognition.

In recent years, convolutional neural networks (CNNs) have become the feature extractors of choice for scene recognition. The previous success of sophisticated pooling leads many studies utilizing CNNs as local features. Early methods adopted a BoF-like approaches, based on the extraction of features from intermediate CNN layers, which were then fed to dictionary learning methods such as clustering \cite{gong2014multi} or sparse coding \cite{dixit2015scene} and pooled by VLAD \cite{gong2014multi} or Fisher vector \cite{liu2014encoding}. In \cite{liu2014encoding}, semantic Fisher vector was proposed, converting features from probability space to the natural parameter space. In \cite{li2017deep}, mixture of factor analyzers Fisher vector was proposed. However, these methods suffer from two drawbacks: 1) the Fisher vector structure is not easy to integrate in CNN, and 2) they are too high-dimensional. These drawbacks prevent end-to-end training and thus leads sub-optimal problem. Recently, VLAD and Fisher vectors are embedded into CNN architecture, by deriving a neural network implementation of its equations. \cite{arandjelovic2016netvlad} proposed NetVLAD, an embedded implementation of VLAD descriptor, and \cite{tang2016deep} proposed Deep FisherNet, an embedded implementation of GMM Fisher vector.

CNNs trained with the ImageNet \cite{donahue2014decaf} for scene recognition was difficult to yield a better result than hand-designed features incorporating with sophisticated classifer \cite{sanchez2013image}. This can be ascribed to the fact that scehe has very distinct characteristics from object classification data. To overcome this problem, \cite{zhou2014learning, zhou2017places} trained a scene-centric CNN by constructing large scale scene dataset, called Places, resulting a significant performance improvement.

In real-world applications, scene images are frequently taken under very different imaging conditions, sensor specifications, and weathers. In such a cross-domain setting, common scene recognition algorithms frequently fail to achieve superior performance. To address the dataset bias problem, many domain adaptation approaches \cite{bruzzone2010domain, duan2012domain, baktashmotlagh2013unsupervised} have been proposed to reduce the mismatch between the data distributions of the training samples and target samples. In \cite{george2016semantic}, semantic clustering (SC), as domain generalization method\footnote{Unlike domain adoptation, in domain generalization, the knowledge learnt from one or multiple source domains in transferred to an unseen target domain.}, for fine-grained scene recognition was proposed.

%% file: Chapter3/chapter3.tex

\chapter{Local Area Transform (LAT)}\label{chap3}

\section{Definition of LAT}
\label{sec:local area transform}
In this paper, we propose to use \emph{local area information} as a robust index for a nonlinear intensity deformation. Let $\bf{I}$ be input image, $\bf{p}$ be the current pixel, $\bf{q}$ be a neighboring pixel, and $\mathcal N_{\bf{p}}$ be a set of neighboring pixels. When denoting a set of pixels $\Psi$ whose intensity value is similar as that of $\bf{p}$ such that $\Psi  = \{ {\bf{\hat q}}|{\bf{I}}({\bf{\hat q}}) \approx {\bf{I}}({\bf{p}}) , {\bf{\hat q}} \in \mathcal N_{\bf{p}}\}$ where $\approx$ means that they have similar values, the local area is defined as the area of $\Psi$. We define a mapping of an image from the intensity domain to the local area domain as local area transform (LAT). LAT is designed to address the matching-problem of a non-linearly deformed image-pair which might be acquired from different radiometric parameters, different photometric parameters, and different modalities (including different spectrums). The LAT at a pixel $\bf{p}$, $\mathcal A(\bf{p})$, is computed as follows:

\begin{equation}
\mathcal A({\bf{\vec p}}) = \sum\limits_{{\bf{q}} \in \mathcal N_{\bf{p}}} {\tau \left( {{\bf{I}}({\bf{p}}),{\bf{I}}({\bf{q}})} \right)},
\end{equation}
where $\tau (x,y) = \left\{ {\begin{array}{*{20}{c}} s(x,y)\\ 0 \end{array}} \right.\begin{array}{*{20}{c}} {\;\;\;\;if\;s(x,y) < thr}\\ {\;else\;\;\;\;} \end{array}$ is a logistic function with definition of similarity function $s$. $s$ is modeled according to the usages and applications. For example, $s$ can be measured as equality check, similarity in spatial domain, similarity in intensity domain, or similarity in gradient domain. When $s$ is modeled as equality check function, $\tau (x,y)$ is defined as a logistic function $\tau (x,y) = \left\{ {\begin{array}{*{20}{c}} 1\\ 0 \end{array}} \right.\begin{array}{*{20}{c}} {\;\;\;\;if\;x = y}\\ {\;else\;\;\;\;} \end{array}$ with $\textstyle {property 1}$: $\tau (kx,ky) = \tau (x,y)$ where $k\in \mathbb N,\;k \ne 0$ and $\textstyle {property 2}$: $\tau (x_1,y_1) = 1$ and $\tau (x_2,y_2) = 1$ $\Rightarrow$ $\tau (x_1 x_2,y_1 y_2) = 1$.

\section{Properties of LAT}
\label{sec:local area transform}
Variety of real world computer vision applications require invariance properties, especially in uncontrolled environments. This section derives the invariance of LAT to non-linear intensity deformations, especially radiometric, photometric, and spectral deformations.
\subsection{Invariance to non-linear intensity deformation}
For a registered input image pair ${\bf{I}}_1$ and ${\bf{I}}_2$, a non-linear intensity deformation between ${\bf{I}}_1$ and ${\bf{I}}_2$ can be represented as $\mathcal D\{ {{\bf{I}}_2}({\bf{p}})\}  = c({{\bf{I}}_1}({\bf{p}})){{\bf{I}}_1}({\bf{p}})$ where $c(\cdot)$ is a intensity mapping operator. Then, $\mathcal A_2(\bf{p})$ is written as follows:

\begin{equation}\begin{array}{l}
{\mathcal A_2}({\bf{p}}) = \sum\limits_{{\bf{q}} \in \mathcal N_{\bf{p}}} {\tau ({{\bf{I}}_2}({\bf{p}}),{{\bf{I}}_2}({\bf{q}}))} \\
\quad \quad \;\; = \sum\limits_{{\bf{q}} \in \mathcal N_{\bf{p}}} {\tau (c({{\bf{I}}_1}({\bf{p}})){{\bf{I}}_1}({\bf{p}}),c({{\bf{I}}_1}({\bf{q}})){{\bf{I}}_1}({\bf{q}}))} \\
\quad \quad \;\; = \sum\limits_{{\bf{q}} \in \mathcal N_{\bf{p}}} {\tau (m{{\bf{I}}_1}({\bf{p}}),n{{\bf{I}}_1}({\bf{q}}))},
\end{array}
\end{equation}
where $m$ and $n$ are constant values varied according to ${\bf{I}}_1(\bf{p})$ and ${\bf{I}}_1(\bf{q})$. For the case of ${\bf{I}}_1(\bf{p}) = {\bf{I}}_1(\bf{q})$ and consequently $m=n$, with $\textstyle{property1}$ of the function $\tau$, ${\tau (m{{\bf{I}}_1}({\bf{p}}),n{{\bf{I}}_1}({\bf{q}}))} = {\tau ({{\bf{I}}_1}({\bf{p}}),{{\bf{I}}_1}({\bf{q}}))}$. For the case of ${\bf{I}}_1(\bf{p}) \ne {\bf{I}}_1(\bf{q})$ and consequently $m \ne n$, under the assumption that the deformation function $\mathcal D$ is an one-to-one mapping, $\tau(m{\bf{I}}_1({\bf{p}}),n{\bf{I}}_1({\bf{q}}))$ is also equal to $\tau({\bf{I}}_1(\bf{p}),{\bf{I}}_1(\bf{q}))$. From these equalities, ${\mathcal A_2}({\bf{p}}) = \sum\limits_{{\bf{q}} \in N_{\bf{p}}} {\tau ({{\bf{I}}_1}({\bf{p}}),{{\bf{I}}_1}({\bf{q}}))} = {\mathcal A_1}({\bf{p}})$. In other words, LAT is invariant to non-linear intensity deformations.

\subsection{Invariance to radiometric \& photometric deformations}
Substituting (3) into (8), $\mathcal A({\bf{p}})$ is rewritten as follows:
\begin{equation}
\mathcal A({\bf{p}}) = \sum\limits_{{\bf{q}} \in {\bf{\mathcal N_{\bf{p}}}}} {\tau \left( {\varepsilon E(T,\lambda )m({\bf{p}})S({\bf{p}},\lambda )v\,,\varepsilon E({T},{\lambda })m({\bf{q}})S({\bf{q}},\lambda )v} \right)}.
\end{equation}

\begin{figure}[!t]
\renewcommand{\thesubfigure}{}
\centering
\includegraphics[width=0.24\linewidth]{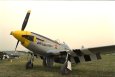}
\includegraphics[width=0.24\linewidth]{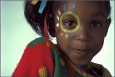}
\includegraphics[width=0.24\linewidth]{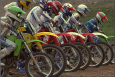}
\includegraphics[width=0.24\linewidth]{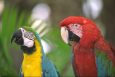}
\linebreak
\linebreak
\includegraphics[width=0.24\linewidth]{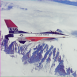}
\includegraphics[width=0.24\linewidth]{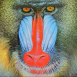}
\includegraphics[width=0.24\linewidth]{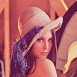}
\includegraphics[width=0.24\linewidth]{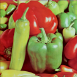}
\caption[The original test color images]{The original test color images used for robustness evaluation and simulated feature matching}
\end{figure}

Under the assumption of local-constancy of $E(T,\lambda)$ and the fact that $\varepsilon$ and $v$ are constant values, (10) is simplified with $\textstyle{property1}$ of the function $\tau$ as:
\begin{equation}
\mathcal A({\bf{p}}) = \sum\limits_{{\bf{q}} \in {\bf{\mathcal N_{\bf{p}}}}} {\tau \left( {m({\bf{p}})S({\bf{p}},\lambda ),m({\bf{q}})S({\bf{q}},\lambda )} \right)}.
\end{equation}

(11) indicates that LAT is independent of the illumination spectral distribution $E(T,\lambda)$ and the exposure time $\varepsilon$, i.e., it is invariant to illumination and exposure deformations (corresponding to radiometric and photometric deformations, respectively).

\subsection{Invariance to spectral deformation}
When we let $\mathcal A^{\lambda_i}(\bf{p})$ be a LATransformed value of an image captured by $i^{th}$-sensor with ${\lambda_i}$ (e.g., visible spectrum) and $\mathcal A^{\lambda_j}(\bf{p})$ be a LATransformed value of an image captured $j^{th}$-sensor with ${\lambda_j}$ (e.g., infra-red spectrum), we show that $\mathcal A^{\lambda_i}({\bf{p}}) = \mathcal A^{\lambda_j}({\bf{p}})$, i.e., the invariance of LAT to a spectral deformation as follows. From (11) $\mathcal A^{\lambda_i}(\bf{p})$ and $\mathcal A^{\lambda_j}(\bf{p})$ are written as (12) and (13), respectively.

\begin{equation}
{\mathcal A^{\lambda_i}}({\bf{p}}) = \sum\limits_{{\bf{q}} \in {\bf{\mathcal N_{\bf{p}}}}} {\tau \left( {{{m(\bf{p})} S({\bf{p}},{\lambda _i}),{m(\bf{q})} S({\bf{q}},{\lambda _i})}} \right)}.
\end{equation}

\begin{equation}
{\mathcal A^{\lambda_j}}({\bf{p}}) = \sum\limits_{{\bf{q}} \in {\bf{\mathcal N_{\bf{p}}}}} {\tau \left( {{{m(\bf{p})} S({\bf{p}},{\lambda _j}),{m(\bf{q})} S({\bf{q}},{\lambda _j})}} \right)}.
\end{equation}

We assume that pixels having same spectral reflectance values for a specific wavelength have same spectral reflectance values for another wavelength, i.e., $\forall {\bf{p}} \ne {\bf{q}} \;\;{\rm{if}}\;S({\bf{p}},{\lambda _i}) = S({\bf{q}},{\lambda _i}),\;\; S({\bf{p}},{\lambda _j}) = S({\bf{q}},{\lambda _j})$. Under this assumption and the $\textstyle{property2}$ of the function $\tau$, ${\tau \left( {{{m(\bf{p})} S({\bf{p}},{\lambda _i}),{m(\bf{q})} S({\bf{q}},{\lambda _i})}} \right)} = {\tau \left( {{{m(\bf{p})} S({\bf{p}},{\lambda _j}),{m(\bf{q})} S({\bf{q}},{\lambda _j})}} \right)}$ when ${m(\bf{p})} = {m(\bf{q})}$. For the case of ${m(\bf{p})} \ne {m(\bf{q})}$, ${\tau \left( {{{m(\bf{p})} S({\bf{p}},{\lambda}),{m(\bf{q})} S({\bf{q}},{\lambda})}} \right)}$ is commonly $0$ except for $\forall m({\bf{p}}) \ne m({\bf{q}})$ and $S({\bf{p}},\lambda) \ne S({\bf{q}},\lambda), {{m(\bf{p})} S({\bf{p}},{\lambda}) = {m(\bf{q})} S({\bf{q}},{\lambda})}$. Note that this exceptional case is out-of consideration since it hardly occurs. Accordingly, ${\mathcal A^{\lambda_i}}({\bf{p}}) = {\mathcal A^{\lambda_j}}({\bf{p}})$ for any wavelength pair $\lambda_i$ and $\lambda_j$, i.e., a LAT value is invariant to a spectral deformation.

\begin{table}[!t]
\caption*{Algorithm 3.1 Pseudo code for LAT}
\centering
\begin{tabular}{l}
\hline
\textbf{Algorithm 1}: Local Area Transform\\
\hline
\textbf{Input:} input image ${\bf{I}}$\\
\textbf{Internal:} Integral histogram ${\bf{H_I}}$, \\ Local histogram $\bf{H_p}$ at pixel point ${\bf{p}}=(x,y)$,\\
the corresponding intensity bin $b$ of ${\bf{p}}$, half-window size $l$\\
\textbf{Output:} Local area transformed image ${\bf{\mathcal A}}$\\
/* integral histogram computation  */ \\
\textbf{for each} pixel (x,y) \textbf{do}\\
${\qquad\bf{H}}'(x,y) \leftarrow {\bf{H}}'(x,y-1)+{\bf{I}}(x,y)$\\
\textbf{end}\\
\textbf{for each} pixel (x,y) \textbf{do}\\
${\qquad\bf{H_I}}(x,y) \leftarrow {\bf{H_I}}(x-1,y)+{\bf{H}}'(x,y)$\\
\textbf{end} \\
/* local histogram computation */ \\
\textbf{for each} pixel (x,y) \textbf{do}\\
$\begin{array}{l}
{\bf{H_p}}(x,y) \leftarrow {\bf{H_I}}(x + l,y + l) + {\bf{H_I}}(x - l,y - l)\\
\quad \quad \quad \quad  - {\bf{H_I}}(x - l,y + l) - {\bf{H_I}}(x + l,y - l)
\end{array}$\\
\textbf{end} \\
/* local area computation */ \\
\textbf{for each} pixel (x,y) \textbf{do}\\
${\qquad\bf{\mathcal A}}(x,y) \leftarrow$ $\sum\limits_{b \in neighbor\,bins} {\omega (b) \times {\bf{H_p}}(x,y,b)}$\\
\textbf{end}\\
\hline
\end{tabular}
\end{table}

\subsection{Limitation}
In the above, we show the invariance of LAT to non-linear intensity deformations. However, when the deformation function $\mathcal D$ is not a one-to-one mapping, there is possibly a duplicated mapping, i.e., $\forall {\bf{I}}_1(\bf{p}) \ne {\bf{I}}_1(\bf{p})$ and consequently $m \ne n$, ${m\bf{I}}_1({\bf{p}}) = {n{\bf{I}}}_1({\bf{p}})$ in (9). For such a mapping, $\mathcal A_1(\bf{p}) \ne \mathcal A_2(\bf{p})$. In other words, the LAT is not invariant to a duplicated intensity deformation. Nevertheless, LAT is still a robust transform to non-linear deformations since non-duplicated deformation assumption is commonly insured.

\begin{figure}[!t]
\renewcommand{\thesubfigure}{}
\centering{
\includegraphics[width=0.19\linewidth]{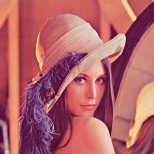}
\includegraphics[width=0.19\linewidth]{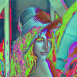}
\includegraphics[width=0.19\linewidth]{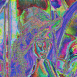}
\includegraphics[width=0.19\linewidth]{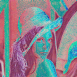}
\includegraphics[width=0.19\linewidth]{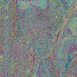}

\includegraphics[width=0.19\linewidth]{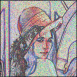}
\includegraphics[width=0.19\linewidth]{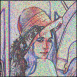}
\includegraphics[width=0.19\linewidth]{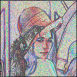}
\includegraphics[width=0.19\linewidth]{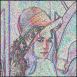}
\includegraphics[width=0.19\linewidth]{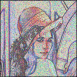}}
\caption[Nonlinear intensity deformation robustness of LAT.]{Nonlinear intensity deformation robustness of LAT. For each sub-figure, top rows are input images and bottom rows are LATransformed images. From left to right, figures are deformed with piecewise linear (PL), piecewise quadratic (PQ), random mapping with Gaussian distribution (RG), and random mapping with uniform distribution (RU).}
\end{figure}

\section{Implementation of LAT and Extension}
The LAT is efficiently computed from a local histogram as $\mathcal A({\bf{p}})=\bf{H_p}({\bf{I}}({\bf{p}}))$. $\bf{H}$ is a $B$-dimensional vector defined as:
\begin{equation}
{{\bf{H}}_{\bf{p}}}(b) = \sum\limits_{{\bf{q}} \in \mathcal N_{\bf{p}}} {{\mathop{\rm Q}\nolimits} ({\bf{I}}({\bf{q}}),b)} ,\quad b \in (1\,,\,B),
\end{equation}
where ${\bf{H}}_{\bf{p}}(b)$ represents the histogram value corresponding to a bin $b$, $B$ is the number of bins, and ${{\mathop{\rm Q}\nolimits}({\bf{I}}({\bf{q}}),b)}$ is zero except when intensity value ${\bf{I}}({\bf{q}})$ belongs to to bin $b$. The computational complexity of the brute-force implementation of the local histograms is linear in the neighboring size. This dependency can be removed using integral histogram \cite{IntegralHistogram} in a way similar to integral image, which reduces the computational complexity from $O(\left|\mathcal N_{\bf{p}}\right|B)$ to $O(B)$ at each pixel location.

For practical usefulness and noise robustness, we employ Gaussian integrated similarity function in intensity domain $s$ instead of the naive definition (with equality check similarity function) $\mathcal A({\bf{p}})=\bf{H_p}({\bf{I}}({\bf{p}}))$ for computing the local area value. Specifically, the local area value is computed by a weighed integration of adjacent bins as (3.8).
\begin{equation}
\mathcal A({\bf{p}}) = K_h\sum\limits_{b \in ({R_k}\,,\,{R_l})} {\omega (b){\bf{H}_p}(b)},
\end{equation}
\begin{equation}
\omega (b) = {e^{ - \frac{{{{\left| {b - {\bf{I}}({\bf{p}})} \right|}^2}}}{{{\sigma ^2}}}}},
\end{equation}
where $K_h = 1/\sum\limits_{b \in ({R_k}{\mkern 1mu} ,{\mkern 1mu} {R_l})} {\omega (b)}$ is a normalization factor, $\omega (b)$ is Gaussian similarity weights of adjacent bins, ${R_k} = {\bf{I}}({\bf{p}}) - r\,$, and ${R_l} = {\bf{I}}({\bf{p}}) + r\,$. Parameters $r$ and $\sigma$ control the interval of integration and the degree of Gaussian smoothing of histogram, respectively. Pseudo code is given in Algorithm 3.1. First, integral Histogram $\bf{H_I}$ is computed through the image, and then local histogram $\bf{H_p}$ at pixel $\bf{p}$ is computed. Lastly, Local area value is computed with Gaussian similarity weights $\omega (b)$. For multiple channel of sensors, e.g., RGB sensor, local area values are computed for each channel, respectively.\\

\section{Robustness Evaluation of LAT}

A non-linear intensity deformation is commonly induced by different modality of imaging system. In order to evaluate the robustness of LAT to non-linear intensity deformation, a challenging simulated database is constructed. Eight color images (Airplane, Baboon, Bikes, Lena, Mustang, PaintedFace, Peppers, TwoMacaws, shown in Fig. 3.1) were employed as original images. Each image is deformed using 40 intensity deformation functions constructed by four categories of random probability distribution: piecewise linear mapping (PL), piecewise quadratic mapping (PQ), random mapping with Gaussian distribution (RG), and random mapping with uniform distribution (RU). For each R, G, B channel different deformation functions were applied. In total, 320 non-linear deformed pairs of color images were generated.


\begin{figure}[!t]
\renewcommand{\thesubfigure}{}
\centering
	\begin{subfigure}[t]{\linewidth}
    	\includegraphics[width=0.19\linewidth]{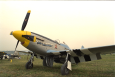}
        \includegraphics[width=0.19\linewidth]{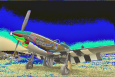}
        \includegraphics[width=0.19\linewidth]{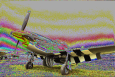}
        \includegraphics[width=0.19\linewidth]{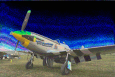}
        \includegraphics[width=0.19\linewidth]{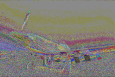}
    \caption{ORG}\label{fig:2a}
    \end{subfigure}    
	\begin{subfigure}[t]{\linewidth}
    	\includegraphics[width=0.19\linewidth]{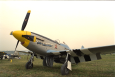}
        \includegraphics[width=0.19\linewidth]{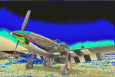}
        \includegraphics[width=0.19\linewidth]{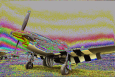}
        \includegraphics[width=0.19\linewidth]{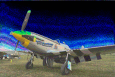}
        \includegraphics[width=0.19\linewidth]{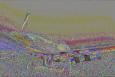}
        \caption{GW}\label{fig:2b}
    \end{subfigure}    
    \begin{subfigure}[t]{\linewidth}
    	\includegraphics[width=0.19\linewidth]{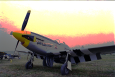}
        \includegraphics[width=0.19\linewidth]{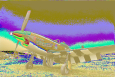}
        \includegraphics[width=0.19\linewidth]{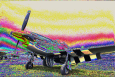}
        \includegraphics[width=0.19\linewidth]{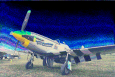}
        \includegraphics[width=0.19\linewidth]{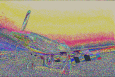}
        \caption{HM}\label{fig:2c}
    \end{subfigure}    
    \begin{subfigure}[t]{\linewidth}
    	\includegraphics[width=0.19\linewidth]{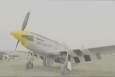}
        \includegraphics[width=0.19\linewidth]{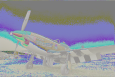}
        \includegraphics[width=0.19\linewidth]{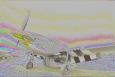}
        \includegraphics[width=0.19\linewidth]{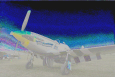}
        \includegraphics[width=0.19\linewidth]{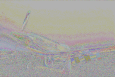}
        \caption{LC}\label{fig:2d}
    \end{subfigure}    
    \begin{subfigure}[t]{\linewidth}
        \includegraphics[width=0.19\linewidth]{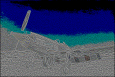}
        \includegraphics[width=0.19\linewidth]{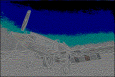}
        \includegraphics[width=0.19\linewidth]{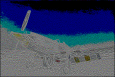}
        \includegraphics[width=0.19\linewidth]{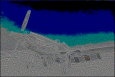}
        \includegraphics[width=0.19\linewidth]{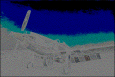}
        \caption{RT}\label{fig:2e}
    \end{subfigure}    
    \begin{subfigure}[t]{\linewidth}
        \includegraphics[width=0.19\linewidth]{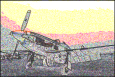}
        \includegraphics[width=0.19\linewidth]{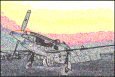}
        \includegraphics[width=0.19\linewidth]{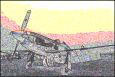}
        \includegraphics[width=0.19\linewidth]{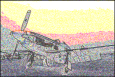}
        \includegraphics[width=0.19\linewidth]{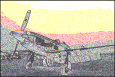}
        \caption{LAT}\label{fig:2f}
    \end{subfigure}    
    \caption[The robustness comparison for nonlinear intensity deformations for Mustang.]{The robustness comparison for nonlinear intensity deformations for Mustang. For each sub-figure, from left to right, figures are non-deformed image, deformed images by piecewise-linear mapping, deformed images by piecewise-quadratic mapping, deformed images by random mapping with Gaussian distribution, deformed images by random mapping with uniform distribution. From top to bottom, original images (ORG), transformed images with grey-world (GW), histogram matching (HM), log-chromaticity (LC), rank transform (RT), and local area transform (LAT, ours). The figures are best viewed in color.}\label{fig:2}
\end{figure}

\begin{figure}[!t]
\renewcommand{\thesubfigure}{}
	\begin{subfigure}[t]{\linewidth}
    \centering
    	\includegraphics[width=0.145\linewidth]{Figures_Supplementary/Fig1_Airplane.png}
        \includegraphics[width=0.145\linewidth]{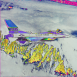}
        \includegraphics[width=0.145\linewidth]{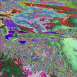}
        \includegraphics[width=0.145\linewidth]{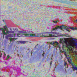}
        \includegraphics[width=0.145\linewidth]{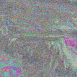}
        \caption{ORG}\label{fig:2a}
    \end{subfigure}
	\begin{subfigure}[t]{\linewidth}
    \centering
    	\includegraphics[width=0.145\linewidth]{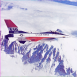}
        \includegraphics[width=0.145\linewidth]{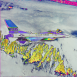}
        \includegraphics[width=0.145\linewidth]{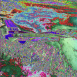}
        \includegraphics[width=0.145\linewidth]{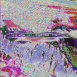}
        \includegraphics[width=0.145\linewidth]{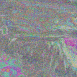}
        \caption{GW}\label{fig:2b}
    \end{subfigure}    
    \begin{subfigure}[t]{\linewidth}
    \centering
    	\includegraphics[width=0.145\linewidth]{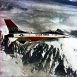}
        \includegraphics[width=0.145\linewidth]{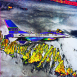}
        \includegraphics[width=0.145\linewidth]{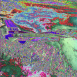}
        \includegraphics[width=0.145\linewidth]{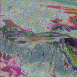}
        \includegraphics[width=0.145\linewidth]{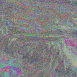}
        \caption{HM}\label{fig:2c}
    \end{subfigure}    
    \begin{subfigure}[t]{\linewidth}
    \centering
    	\includegraphics[width=0.145\linewidth]{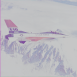}
        \includegraphics[width=0.145\linewidth]{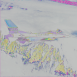}
        \includegraphics[width=0.145\linewidth]{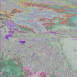}
        \includegraphics[width=0.145\linewidth]{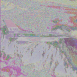}
        \includegraphics[width=0.145\linewidth]{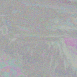}
        \caption{LC}\label{fig:2d}
    \end{subfigure}    
    \begin{subfigure}[t]{\linewidth}
    \centering
        \includegraphics[width=0.145\linewidth]{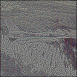}
        \includegraphics[width=0.145\linewidth]{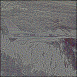}
        \includegraphics[width=0.145\linewidth]{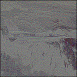}
        \includegraphics[width=0.145\linewidth]{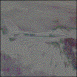}
        \includegraphics[width=0.145\linewidth]{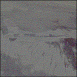}
        \caption{RT}\label{fig:2e}
    \end{subfigure}    
    \begin{subfigure}[t]{\linewidth}
    \centering
        \includegraphics[width=0.145\linewidth]{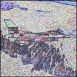}
        \includegraphics[width=0.145\linewidth]{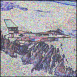}
        \includegraphics[width=0.145\linewidth]{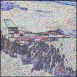}
        \includegraphics[width=0.145\linewidth]{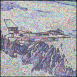}
        \includegraphics[width=0.145\linewidth]{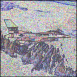}
        \caption{LAT}\label{fig:2f}
    \end{subfigure}  
    
    \caption[The robustness comparison for nonlinear intensity deformations for Airplane.]{The robustness comparison for nonlinear intensity deformations for Airplane. For each sub-figure, from left to right, figures are non-deformed image, deformed images by piecewise-linear mapping, deformed images by piecewise-quadratic mapping, deformed images by random mapping with Gaussian distribution, deformed images by random mapping with uniform distribution. From top to bottom, original images (ORG), transformed images with grey-world (GW), histogram matching (HM), log-chromaticity (LC), rank transform (RT), and local area transform (LAT, ours). The figures are best viewed in color.}\label{fig:2}
\end{figure}

\begin{figure}[!t]
\renewcommand{\thesubfigure}{}
\centering{
	\begin{subfigure}[t]{\linewidth}
    \centering
    	\includegraphics[width=0.145\linewidth]{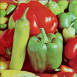}
        \includegraphics[width=0.145\linewidth]{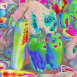}
        \includegraphics[width=0.145\linewidth]{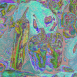}
        \includegraphics[width=0.145\linewidth]{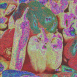}
        \includegraphics[width=0.145\linewidth]{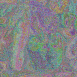}
        \caption{ORG}\label{fig:2a}
    \end{subfigure}    
	\begin{subfigure}[t]{\linewidth}
    \centering
    	\includegraphics[width=0.145\linewidth]{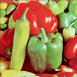}
        \includegraphics[width=0.145\linewidth]{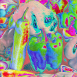}
        \includegraphics[width=0.145\linewidth]{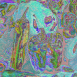}
        \includegraphics[width=0.145\linewidth]{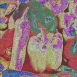}
        \includegraphics[width=0.145\linewidth]{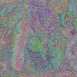}
        \caption{GW}\label{fig:2b}
    \end{subfigure}    
    \begin{subfigure}[t]{\linewidth}
    \centering
    	\includegraphics[width=0.145\linewidth]{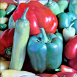}
        \includegraphics[width=0.145\linewidth]{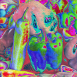}
        \includegraphics[width=0.145\linewidth]{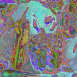}
        \includegraphics[width=0.145\linewidth]{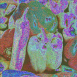}
        \includegraphics[width=0.145\linewidth]{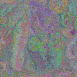}
        \caption{HM}\label{fig:2c}
    \end{subfigure}    
    \begin{subfigure}[t]{\linewidth}
    \centering
    	\includegraphics[width=0.145\linewidth]{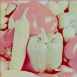}
        \includegraphics[width=0.145\linewidth]{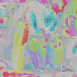}
        \includegraphics[width=0.145\linewidth]{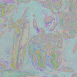}
        \includegraphics[width=0.145\linewidth]{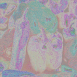}
        \includegraphics[width=0.145\linewidth]{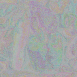}
        \caption{LC}\label{fig:2d}
    \end{subfigure}    
    \begin{subfigure}[t]{\linewidth}
    \centering
        \includegraphics[width=0.145\linewidth]{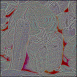}
        \includegraphics[width=0.145\linewidth]{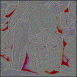}
        \includegraphics[width=0.145\linewidth]{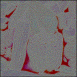}
        \includegraphics[width=0.145\linewidth]{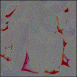}
        \includegraphics[width=0.145\linewidth]{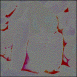}
        \caption{RT}\label{fig:2e}
    \end{subfigure}    
    \begin{subfigure}[t]{\linewidth}
    \centering
        \includegraphics[width=0.145\linewidth]{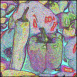}
        \includegraphics[width=0.145\linewidth]{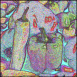}
        \includegraphics[width=0.145\linewidth]{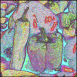}
        \includegraphics[width=0.145\linewidth]{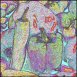}
        \includegraphics[width=0.145\linewidth]{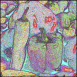}
        \caption{LAT}\label{fig:2f}
    \end{subfigure}    
    \caption[The robustness comparison for nonlinear intensity deformations for Pepper.]{The robustness comparison for nonlinear intensity deformations for Pepper. For each sub-figure, from left to right, figures are non-deformed image, deformed images by piecewise-linear mapping, deformed images by piecewise-quadratic mapping, deformed images by random mapping with Gaussian distribution, deformed images by random mapping with uniform distribution. From top to bottom, original images (ORG), transformed images with grey-world (GW), histogram matching (HM), log-chromaticity (LC), rank transform (RT), and local area transform (LAT, ours). The figures are best viewed in color.}\label{fig:2}
    }
\end{figure}

The robustness of LAT was evaluated with comparisons to four methods: grey-world model (GW) \cite{NormalizedChromaticity}, histogram matching (HM) \cite{HistogramMatching}, log-chromaticity (LC) \cite{ANCC}, and rank transform (RT) \cite{Rank}. As a base, the original image pair before transform (ORG) was also compared.

\begin{table}[!t]
\begin{center}
\caption[Similarity comparison results in terms of mean absolute difference ($d_{mad}$) and different pixel ratio ($d_{dpr}$) for all image pairs.]{Similarity comparison results in terms of mean absolute difference ($d_{mad}$) and different pixel ratio ($d_{dpr}$) for all piecewise linear mapping (PL), piecewise quadratic mapping (PQ), random mapping with Gaussian distribution (RG), and random mapping with uniform distribution (RU) image pairs.}
\centering
\setlength{\extrarowheight}{5.0pt}
\begin{tabular}{c|cccccc}
\hline
Deform & ORG & GW \cite{NormalizedChromaticity} & HM \cite{HistogramMatching} & LC \cite{ANCC} & RT \cite{Rank} & LAT\\
\hline
\hline
$d_{mad}$ & 0.33 & 0.13 & 0.32 & 0.12 & 0.20 & \textbf{0.02} \\
$d_{dpr}$ & 0.79 & 0.53 & 0.78 & 0.46 & 0.53 & \textbf{0.04}\\
\hline
\end{tabular}
\end{center}
\end{table}
\vspace{-2em}

\begin{table}[!t]
\begin{center}
\caption[Similarity comparison results in terms of mean absolute difference ($d_{mad}$) and different pixel ratio ($d_{dpr}$) for non-linear deformation as varying the parameters in LAT.]{Similarity comparison results in terms of mean absolute difference ($d_{mad}$) and different pixel ratio ($d_{dpr}$) for non-linear deformation as varying the parameters in LAT. In each experiment, all other parameters are fixed as initial values in Section 4.1.}
\centering
\setlength{\extrarowheight}{5.0pt}
\begin{tabular}{c|ccc|ccc|ccc}
\hline
~ & \multicolumn{3}{c}{window size $\it{l}$} & \multicolumn{3}{c}{interval of integ. $\it{r}$} & \multicolumn{3}{c}{deg. of Gaussian $\sigma$} \\
~ & 7 & 11 & 15 & 1 & 3 & 5 & 0.1 & 0.3 & 0.5\\
\hline
\hline
$d_{mad}$ & 0.11 & 0.02 & 0.06 & 0.03 & 0.02 & 0.04 & 0.03 & 0.02 & 0.06\\
$d_{dpr}$ & 0.13 & 0.04 & 0.08 & 0.06 & 0.04 & 0.05 & 0.07 & 0.04 & 0.08\\
\hline
\end{tabular}
\end{center}
\end{table}

The similarity between a registered image pair is measured by the mean absolute difference $d_{mad}$ and different pixel ratio $d_{dpr}$. $d_{mad}$ and $d_{dpr}$ are defined as (21) and (22), respectively.
\begin{equation}
d_{mad} = {K_1}\sum\limits_{\bf{p}} {|{{{\bf{\bar I}}}_1}({\bf{p}}) - {{{\bf{\bar I}}}_2}({\bf{p}})|},
\end{equation}
where ${{{\bf{\bar I}}}_1}$ is a transformed version of the original image, ${{{\bf{\bar I}}}_2}$ is a transformed version of the deformed image, $K_1 = 1/(N L)$ is a normalization factor, $N$ is the number of pixels, $L$ is the maximum value of the label.
\begin{equation}
d_{dpr} = {K_2}\sum\limits_{\bf{p}} {(|{{{\bf{\bar I}}}_1}({\bf{p}}) - {{{\bf{\bar I}}}_2}({\bf{p}})|\, > t)},
\end{equation}
where $K_2 = 1/N$ is a normalization factor, $t=0.1L$ is threshold value.

The qualitative evaluations for LAT are summarized in Table 3.1 and Table 3.2, showing that LAT is superior to the other methods in terms of both $d_{mad}$ and $d_{dpr}$. It should be noted that lower $d_{mad}$ and $d_{dpr}$ are, the more similar sample image pairs are. In the results, the total 10 image pairs are used for an average, and sample images are represented in Fig. 3.2 - Fig. 3.5. More specifically an input image is non-linearly transformed with different transformations, and the reconstruction results are represented as varying image transformation methods, including the state of-the-art method and proposed LAT. The LAT transformed non-linearly deformed images into a common domain, where the discrepancy between non-linear deformations are highly reduced. For most image pairs, the LATransformed images are very similar to each other; in other words, LAT shows higher robustness for randomly intensity-deformed image pairs.

Especially, Table 3.2 intensively analyzed the performance of the LAT as varying associated parameters, including support window size $l$, the interval of integration $r$, and degree of Gaussian smoothing $\sigma$. The performance of LAT was the highest when the parameter $l$ was 11. Note that other parameters $r$ and $\sigma$ in LAT, which control the interval of integration and degree of Gaussian smoothing of the histogram, were not seriously effecting on the performances, thus they were set as $r$ = 3 and $\sigma$ = 0.3 for considering the trade-off between efficiency and robustness.

\clearpage

\section{LAT Reformulated Features: Cross-Modality Feature Descriptors}

Besides the use as a transformation, the concept of \emph{Local Area Transform} can be used to reformulate conventional cost functions and descriptors. If we replace an `intensity value' by a `local area value', it endows cost functions and descriptors with robustness to a modality deformation with maintaining inherent properties of them. For example, the most widely used cost function, a mean absolute difference (mad), can be reformulated as follows:

\begin{equation}
mad({\bf{p}},{\bf{q}}) = \sum\limits_{(x,y) \in {\rm{N}}} {\left| {{{\bf{I}}_{\bf{p}}}(x,y) - {{\bf{I}}_{\bf{q}}}(x,y)} \right|}
\end{equation}
\begin{equation}
ma{d_{LAT}}({\bf{p}},{\bf{q}}) = \sum\limits_{(x,y) \in {\rm{N}}} {\left| {{A_{\bf{p}}}(x,y) - {A_{\bf{q}}}(x,y)} \right|}
\end{equation}
where $mad({\bf{p}},{\bf{q}})$ and $ma{d_{LAT}}({\bf{p}},{\bf{q}})$ are original and the reformulated mad between pixel points {\bf{p}} and {\bf{q}}. ${\rm{N}}$ is the neighbor pixels around {\bf{p}} or {\bf{q}}.

Similarly, the original local self-similarity descriptor (LSS) \cite{LSS} can be reformulated by measuring sum of squared local area difference instead of sum of squared intensity difference as follows:

\begin{equation}
S({\bf{p}},{\bf{q}}) = \exp (\frac{{ss{d_{{\bf{pq}}}}(x,y)}}{{{{{\mathop{\rm var}} }_{auto}}}})
\end{equation}
\begin{equation}
ss{d_{{\bf{pq}}}}(x,y) = \sum\limits_{(x,y)} {{{\{ {{\bf{I}}_{\bf{p}}}(x,y) - {{\bf{I}}_{\bf{q}}}(x,y)\} }^2}}
\end{equation}

\begin{equation}
S_{LAT}({\bf{p}},{\bf{q}}) = \exp (\frac{{sa{d_{{\bf{pq}}}}(x,y)}}{{{{{\mathop{\rm var}} }_{auto}}}})
\end{equation}
\begin{equation}
sa{d_{{\bf{pq}}}}(x,y) = \sum\limits_{(x,y)} {{{\{ {\mathcal A_{\bf{p}}}(x,y) - {\mathcal A_{\bf{q}}}(x,y)\} }^2}}
\end{equation}
where $S({\bf{p}},{\bf{q}})$ and $S_{LAT}({\bf{p}},{\bf{q}})$ are the original and the reformulated correlation surface functions in LSS (please refer \cite{LSS} for full description of LSS). ${\mathop{\rm var}}_{auto}$ is a constant for stability. 

SIFT also can be reformulated by using gradients of local area value (3.19) instead of gradients of intensity value (3.18). 

\begin{equation}
\nabla I(x,y) = \left[ {\frac{{\partial I}}{{\partial x}}\quad \frac{{\partial I}}{{\partial y}}} \right]
\end{equation}
\begin{equation}
\nabla \mathcal A(x,y) = \left[ {\frac{{\partial {\mathcal A}}}{{\partial x}}\quad \frac{{\partial {\mathcal A}}}{{\partial y}}} \right]
\end{equation}

Binary pattern based robust descriptors, e.g., CT \cite{BRISK}, RT \cite{Rank}, BRIEF \cite{BRIEF}, and BRISK \cite{BRISK}, are formulated with following local binary pattern (LBP) equation. 

\begin{equation}
LBP({\bf{p}}) = \sum\limits_{q = 0}^{Q - 1} {\Lambda ({I_q} - {I_{\bf{p}}}){2^q}} \quad \quad \Lambda (x) = \left\{ {\begin{array}{*{20}{c}}
{1,}\\
{0,}
\end{array}} \right.\begin{array}{*{20}{c}}
{x \ge 0}\\
{\quad otherwise}
\end{array}
\end{equation}
where $LBP({\bf{p}})$ is LBP at pixel ${\bf{p}}$. $q(=0,1,...,Q-1)$ is the index of neighboring pixels of ${\bf{p}}$. (3.20) can be reformulated to $LB{P_{LAT}}({\bf{p}})$ with local area value instead of intensity value as follows:
\begin{equation}
LB{P_{LAT}}({\bf{p}}) = \sum\limits_{q = 0}^{Q - 1} {\Lambda ({A_q} - {A_{\bf{p}}}){2^q}} \quad \quad \Lambda (x) = \left\{ {\begin{array}{*{20}{c}}
{1,}\\
{0,}
\end{array}} \right.\begin{array}{*{20}{c}}
{x \ge 0}\\
{\quad otherwise}
\end{array}
\end{equation}
With the reformulated LBP, robust descriptors: CT \cite{BRISK}, RT \cite{Rank}, BRIEF \cite{BRIEF}, and BRISK \cite{BRISK} can be reformulated with LAT. We use the subscription $_{LAT}$ as the meaning of the reformation with LAT in the remaining parts of this paper. Note that any cost functions or features computed from intensity values can be reformulated with LAT.

\vspace{1em}
\begin{figure}[!b]
\renewcommand{\thesubfigure}{}
\centering
\includegraphics[width=0.8\linewidth]{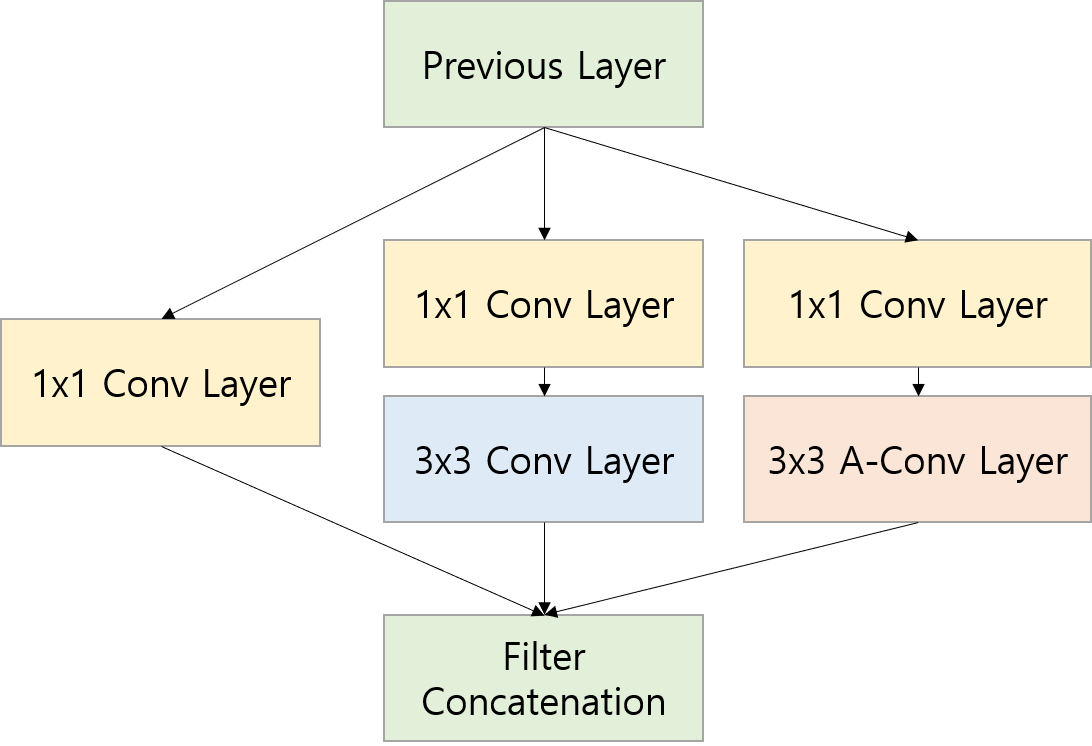}
\caption[Aception block]{Aception block: inception-like stem block composing of concatenated 1x1 Conv, 3x3 Conv, and 3x3 A-Conv layers.}
\end{figure}

\begin{figure}[!t]
\renewcommand{\thesubfigure}{}
\centering
\includegraphics[width=0.4\linewidth]{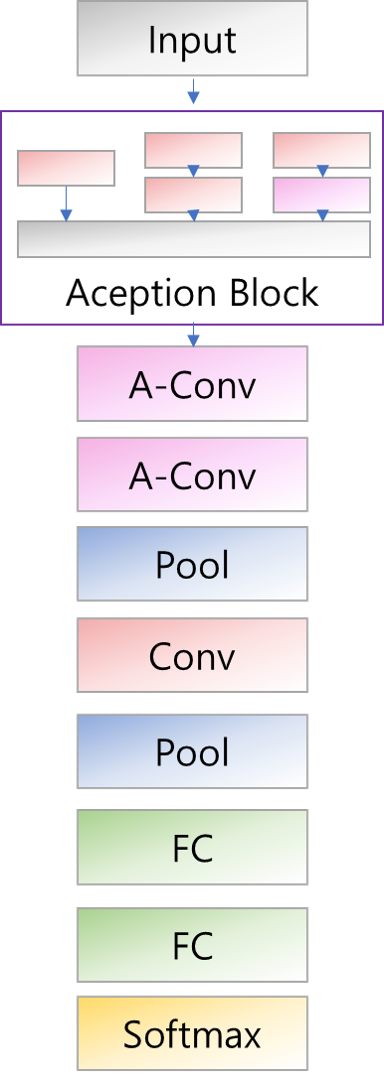}
\caption[The structure of LAT-AlexNet.]{The structure of LAT-AlexNet: Each A-Conv and Conv layers are followed by leaky-ReLu layers and Aception block composing of concatenated 1x1 Conv, 3x3 Conv, and 3x3 A-Conv layers.}
\end{figure}

\section{LAT-Net: Deep Scene Recognition Network}
Scene recognition is one of the fundamental task in various applications of computer vision such as robotics, image search, geo-localization, etc. However, scene recognition is challenging problem since scenes contain variety of components from objects to scene-like features. Furthermore, in practical applications, scene images are frequently taken under cross-domain settings, such as different imaging conditions, sensor specifications, and even weathers. Conventional scene recognition algorithms failed to achieve reliable results. To address this problem, domain adaptation \cite{duan2012domain, baktashmotlagh2013unsupervised} or domain generalization \cite{george2016semantic} approaches have bee proposed. This section proposes to embed LAT concept into deep convolutional neural network (CNNs) in order to tackle cross-domain scene recognition problem.

The conventional convolutional (Conv) layer in common CNNs is defined as: 
\begin{equation}
{{\bf{x}}^l} = \sum\limits_{i \in {\bf{K}}} {{{\bf{\omega}} _i ^{l}}{{\bf{x}}^{l - 1}}_i + b^l}
\end{equation}
where $\bf{x}^l$ and $\bf{x}^{l-1}$ are feature maps of current $l^{th}$ and $l-1^{th}$ layers, respectively. ${\bf{\omega}} _i ^{l}$ and $b^l$ are weights and bias terms. $\bf{K}$ is convolutional kernel. With the concept of LAT, reformulated convolutional (A-Conv) layer is defined as follows:
\begin{equation}
{\mathcal A^l} = \sum\limits_{i \in {\bf{K}}} {{{\bf{\omega}} _i ^{l}}{{\mathcal A}^{l - 1}}_i + b^l}
\end{equation}
where $\mathcal A^l$ and $\mathcal A^{l-1}$ are LAT-reformulated feature maps of current $l^{th}$ and $l-1^{th}$ layers, respectively. $\mathcal A^{l-1}$ could be replaced by the output feature maps of regular layers in a CNN such as Conv layer or a pooling layer. It could be also be a previous A-Conv layer, and thus can be stacked together to form a highly nonlinear transformation operator.

Given the impressive performance on the ImageNet benchmark \cite {krizhevsky2012imagenet, russakovsky2015imagenet}, three popular CNN architectures AlexNet \cite{krizhevsky2012imagenet}, VGG-16 \cite{simonyan2014very}, ResNet-34 \cite{he2016deep} are employed as basis networks. In order to apply non-linear feature transformation into networks, the former Conv layers are replaced as A-Conv layers in the proposed network structures: two Conv layers, four conv layers, six conv layers for AlexNet, VGGNet-16, ResNet-34, respectively. In addition, inception-like stem block, named as Aception block (Fig. 3.6) is placed at the top of each networks. The re-designed CNNs are named as LAT-CNN, i.e., LAT-AlexNet, LAT-VGGNet, and LAT-ResNet, respectively. The structures of re-designed networks are depicted in Figs. 3.7, 3.8, and 3.9, respectively. All the CNNs presented here were implemented and trained using Caffe package \cite{jia2014caffe} on Nvidia GPUs Tesla P40.

\begin{figure}[!t]
\renewcommand{\thesubfigure}{}
\centering
\includegraphics[width=0.22\linewidth]{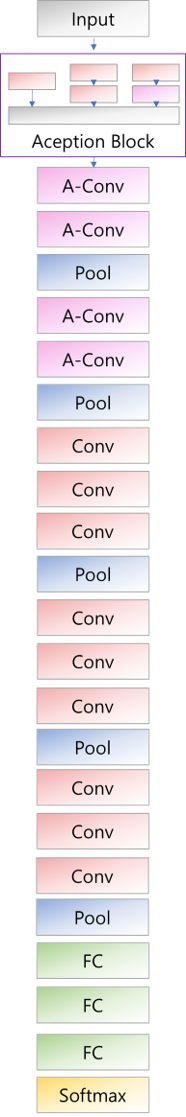}
\caption[The structure of LAT-VGGNet.]{The structure of LAT-VGGNet: Each A-Conv and Conv layers are followed by leaky-ReLu layers and Aception block composing of concatenated 1x1 Conv, 3x3 Conv, and 3x3 A-Conv layers.}
\end{figure}

\begin{figure}[!t]
\renewcommand{\thesubfigure}{}
\centering
\includegraphics[width=0.70\linewidth]{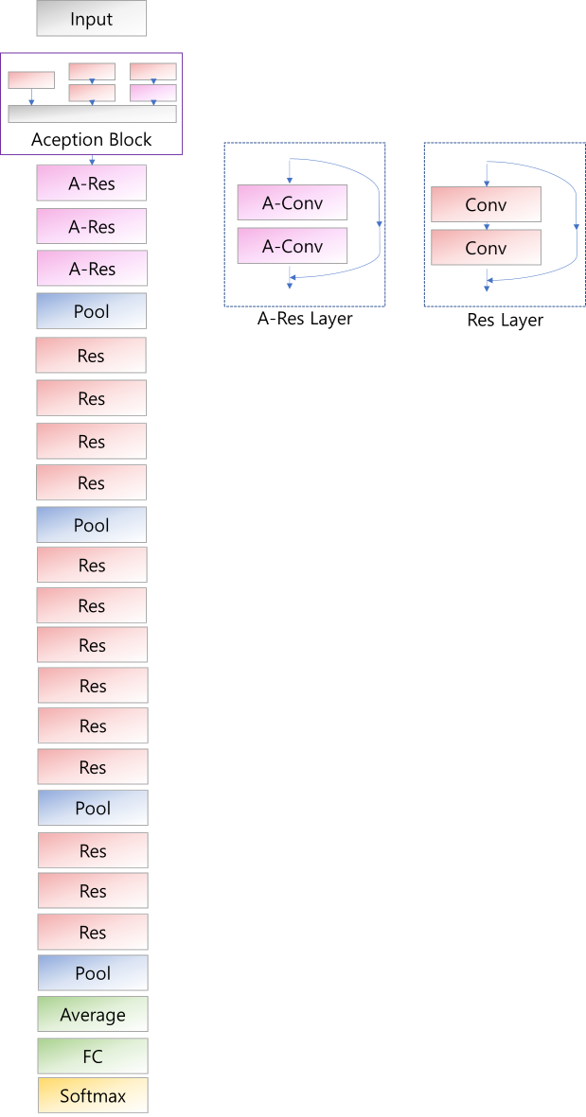}
\caption[The structure of LAT-ResNet.]{The structure of LAT-ResNet: Each A-Conv and Conv layers are followed by leaky-ReLu layers and Aception block composing of concatenated 1x1 Conv, 3x3 Conv, and 3x3 A-Conv layers.}
\end{figure}
\clearpage

%% file: Chapter4/chapter4.tex
\chapter{Cross-Modality Correspondence Matching and Deep Scene Recognition}\label{chap4}

\section{Experimental Settings}
In experiments, the LAT was implemented with the following parameter settings for all datasets: {\it{i}, \it{l}, $\sigma$}={11, 3, 0.3}. LAT was implemented in C++ on Intel Core i7-3770 CPU at 3.40 GHz. In experiments, the performances of LAT were evaluated for the tasks of nonlinearly-deformed image matching in Section 4.2, cross-spectral correspondence matching in Section 4.3, cross-radiometry stereo matching in Section 4.4, and cross-modal dense flow estimation in Section 4.5.  For color images, LAT is computed for each channel, and then those values are used for minimum distance/cost selection. LAT was implemented as C++ layer in deep learning library Caffe \cite{jia2014caffe} for deep scene recognition in Section 4.6.

\section{Cross-Modality Correspondence Matching}
\subsection{Non-linear Deformation Correspondence Matching}
The performance of reformulated feature descriptors with LAT is evaluated in terms of the feature recognition rate. The feature recognition rate is defined as the ratio of corrected matching to the total keypoints similar in \cite{BRIEF}. The keypoints were detected using SIFT detector. SIFT\cite{SIFT}, BRIEF\cite{BRIEF}, and LSS\cite{LSS} were selected as compared feature descriptors since they are the most successful feature descriptors respectively based on gradient, binary pattern, and self-similarity. They are reformulated with LAT to SIFT$_{LAT}$, BRIEF$_{LAT}$, and LSS$_{LAT}$, respectively. For the evaluation, a simulated database described in Section 3.4 were used.

\begin{figure}[!t]
\renewcommand{\thesubfigure}{}
\centering
	\begin{subfigure}[t]{0.49\linewidth}
    	\includegraphics[width=\linewidth]{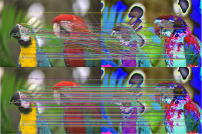}
        \caption{PL}\label{fig:2a}
    \end{subfigure}
    \begin{subfigure}[t]{0.49\linewidth}
    	\includegraphics[width=\linewidth]{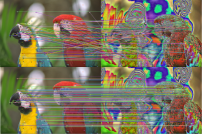}
        \caption{PQ}\label{49:2b}
    \end{subfigure}
    \begin{subfigure}[t]{0.49\linewidth}
    	\includegraphics[width=\linewidth]{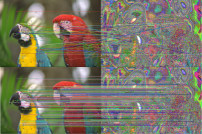}
        \caption{RG}\label{fig:2c}
    \end{subfigure}
    \begin{subfigure}[t]{0.49\linewidth}
    	\includegraphics[width=\linewidth]{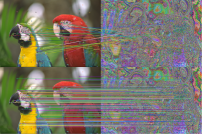}
        \caption{RU}\label{fig:2d}
    \end{subfigure}
    \caption[Feature matching on simulated database.]{Feature matching on simulated database. For each subfigure, upper is a results of SIFT and below is that of SIFT$_{LAT}$. PL: image pairs from piecewise-linear mapping, PQ: image pairs from piecewise-quadratic mapping, RG: image pairs from random mapping with Gaussian distribution, and RU: image pairs from random mapping with uniform distribution.}\label{fig:2}
\end{figure}

Fig. 4.1 shows an example of comparison on a simulated image pair. In the results, the correspondence estimations with conventional SIFT descriptor are represented on the upper part, while that with proposed SIFT descriptor on LAT are represented on the below part. For establishing correspondence, same fixed parameters are used (e.g., same threshold for matching). In other words, the number of correspondence depends on the robustness of the descriptors. In these results, the LAT-based SIFT descriptor provides consistently outperformed correspondences compared to original one. Fig. 4.2 summarizes the overall results representing that the reformulated descriptors remarkably outperforms the original descriptors. Especially, reformulated descriptors shows extremely high recognition rate even for image pairs generated with random mapping function (RG, RU). The results give an insight that the nonlinear intensity deformation problem generally induced by different imaging modalities can be addressed by reformulating the conventional descriptors with LAT. In the remaining parts of this section, we show the superiority and applicability of LAT for several multi-modality applications.

\begin{figure}[!t]
\renewcommand{\thesubfigure}{}
\centering{
\includegraphics[width=0.95\linewidth]{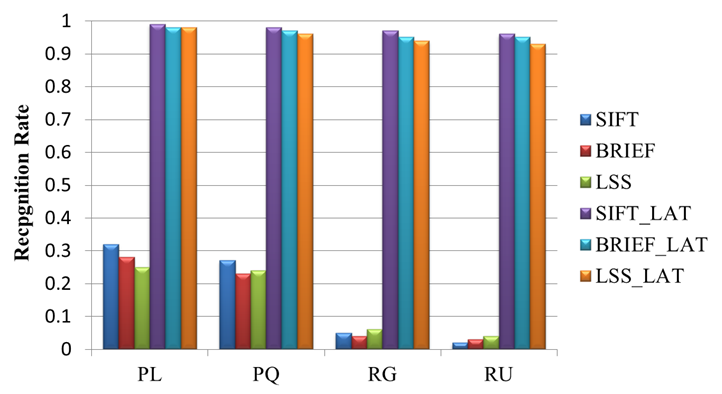}
}
\caption[Recognition rate for simulated database.]{Recognition rate for simulated database. PL: image pairs from piecewise-linear mapping, PQ: image pairs from piecewise-quadratic mapping, RG: image pairs from random mapping with Gaussian distribution, and RU: image pairs from random mapping with uniform distribution.}
\end{figure}

\subsection{Cross-Spectral Correspondence Matching}

In this section, we show that LAT is superior in terms of detecting the sought template in different spectral images, i.e., cross-spectral template matching. The cross-spectral template matching was applied on 100 RGB-NIR image pairs randomly selected from RGB-NIR Scene Dataset \cite{DB1}. For each input NIR image, a template of a give size was selected at 100 random locations. In total, 10,000 (RGB) image-(NIR) template pairs were used in this experiment. To avoid a homogeneous template, the locations of the template were selected from among the structured regions of the image (i.e., locations where the features response of BRISK \cite{BRISK} is above a threshold). Given an RGB image and a NIR-template, matching distances\footnote[1]{The minimum distance among NIR/R-channel, NIR/G-channel, NIR/B-channel distances is set to the distance of the location.} were computed for all possible locations in the corresponding RGB image, and the region associated with the minimal distance was considered the matched region. Four different methods, HM \cite{HistogramMatching}, LC \cite{ANCC}, RT \cite{Rank}, and MTM \cite{MTM} were employed as compared methods and the original images were also compared as a base method. Euclidean distance is employed for ORG, HM, and LC and sum of different rank is employed for RT.

\begin{sidewaysfigure}[!t]
\renewcommand{\thesubfigure}{}
\centering{
\includegraphics[width=0.24\linewidth]{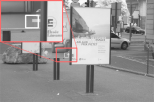}
\includegraphics[width=0.24\linewidth]{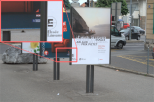}
\includegraphics[width=0.24\linewidth]{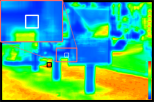}
\includegraphics[width=0.24\linewidth]{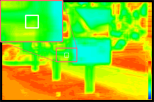}

\includegraphics[width=0.24\linewidth]{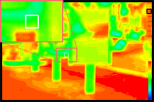}
\includegraphics[width=0.24\linewidth]{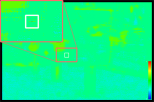}
\includegraphics[width=0.24\linewidth]{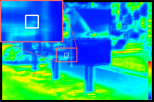}
\includegraphics[width=0.24\linewidth]{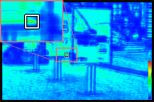}}

\caption[Pixel-wise template response map with a template size of 30.]{Pixel-wise template response map with a template size of 30. For each sub-figure, from left-top to right-bottom, figures are NIR image, RGB image, similarity maps for ORG, HM \cite{HistogramMatching}, LC \cite{ANCC}, RT \cite{Rank}, MTM \cite{MTM} and LAT, respectively. White and black rectangular boxes are a sought template and selected window. A left-top small image is the magnified map around the sought template. Compared to other methods, a sough template candidate is well localized in LATansformed images. The figure is best viewed in color}
\end{sidewaysfigure}

In Fig. 4.3, 4.4, 4.5, in order to evaluate the performance of the LAT, the template matching performances across cross-spectral images are measured compared to the state-of-the-art methods. We show examples of similarity maps (for better visualization, a similarity map, which is the inverse of distance map, is illustrated instead of distance map where higher value (red) means similar region and lower vale (blue) means dissimilar region). The template matching in the LATransformed images clearly shows a sharp peak at the correct location, while it is not well localized in other methods. Table 4.1 summarizes the average correct detection ratio $r_{cd}$ and matching pixel error $e_{mp}$. $r_{cd}$ measures the percentage of correct detection (if matched and true windows are overlapped with $>$70\%, the match is considered a correct detection), and $e_{mp}$ measures the absolute difference between matched and true windows. As shown in Table 4.2, quantitative evaluation of LAT are represented as an average for 10,000 RGB-NIR template pairs LAT provides robust results in cross-spectral template matching in terms of both  $r_{cd}$ and $e_{mp}$; in this study, $r_{cd}$ showed improvement of 23\%, and $e_{mp}$ showed a reduction of 113 pixels.\\

\begin{figure}[!ht]
\renewcommand{\thesubfigure}{}
\centering
	\begin{subfigure}[t]{\linewidth}
    \centering
    	\includegraphics[width=0.25\linewidth]{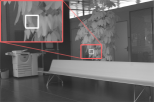}
        \includegraphics[width=0.25\linewidth]{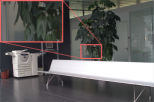}
        \includegraphics[width=0.25\linewidth]{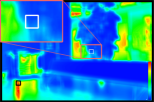}
        \caption{ORG}\label{fig:2a}
    \end{subfigure}
    \begin{subfigure}[t]{\linewidth}
    \centering
    	\includegraphics[width=0.25\linewidth]{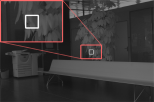}
        \includegraphics[width=0.25\linewidth]{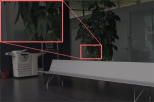}
        \includegraphics[width=0.25\linewidth]{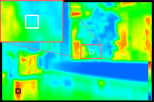}
        \caption{GW}\label{fig:2b}
    \end{subfigure}    
    \begin{subfigure}[t]{\linewidth}
    \centering
    	\includegraphics[width=0.25\linewidth]{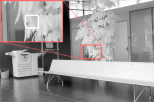}
        \includegraphics[width=0.25\linewidth]{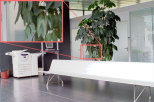}
        \includegraphics[width=0.25\linewidth]{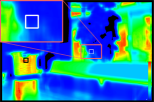}
        \caption{HM}\label{fig:2c}
    \end{subfigure}    
    \begin{subfigure}[t]{\linewidth}
    \centering
    	\includegraphics[width=0.25\linewidth]{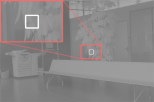}
        \includegraphics[width=0.25\linewidth]{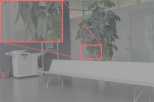}
        \includegraphics[width=0.25\linewidth]{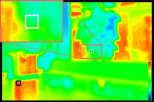}
        \caption{LC}\label{fig:2d}
    \end{subfigure}    
    \begin{subfigure}[t]{\linewidth}
    \centering
    	\includegraphics[width=0.25\linewidth]{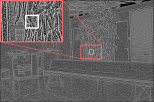}
        \includegraphics[width=0.25\linewidth]{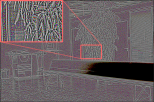}
        \includegraphics[width=0.25\linewidth]{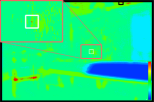}
        \caption{RT}\label{fig:2e}
    \end{subfigure}    
    \begin{subfigure}[t]{\linewidth}
    \centering
    	\includegraphics[width=0.25\linewidth]{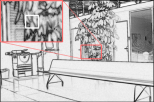}
        \includegraphics[width=0.25\linewidth]{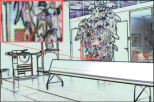}
        \includegraphics[width=0.25\linewidth]{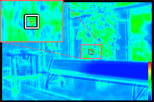}
        \caption{LAT}\label{fig:2f}
    \end{subfigure}    
    \caption[Qualitative results of cross-spectral template matching for Lobby.]{Qualitative results of cross-spectral template matching for Lobby. From left to right transformed NIR image, transformed RGB image, and similarity maps are given. ORG: original NIR and RGB image, HM \cite{HistogramMatching}, LC \cite{ANCC}, RT \cite{Rank}, MTM \cite{MTM} are compared with LAT (ours). White and black rectangular boxes are a sought template and selected window. The figures are best viewed in color.}\label{fig:2}
\end{figure}

\begin{figure}[!ht]
\renewcommand{\thesubfigure}{}
\centering
	\begin{subfigure}[t]{\linewidth}
    \centering
    	\includegraphics[width=0.25\linewidth]{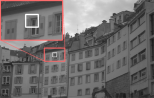}
        \includegraphics[width=0.25\linewidth]{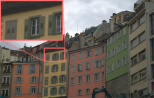}
        \includegraphics[width=0.25\linewidth]{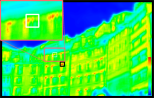}
        \caption{ORG}\label{fig:2a}
    \end{subfigure}
    \begin{subfigure}[t]{\linewidth}
    \centering
    	\includegraphics[width=0.25\linewidth]{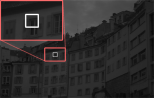}
        \includegraphics[width=0.25\linewidth]{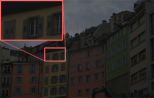}
        \includegraphics[width=0.25\linewidth]{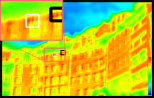}
        \caption{GW}\label{fig:2b}
    \end{subfigure}    
    \begin{subfigure}[t]{\linewidth}
    \centering
    	\includegraphics[width=0.25\linewidth]{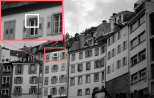}
        \includegraphics[width=0.25\linewidth]{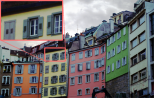}
        \includegraphics[width=0.25\linewidth]{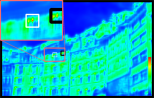}
        \caption{HM}\label{fig:2c}
    \end{subfigure}    
    \begin{subfigure}[t]{\linewidth}
    \centering
    	\includegraphics[width=0.25\linewidth]{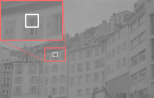}
        \includegraphics[width=0.25\linewidth]{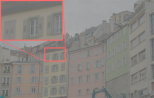}
        \includegraphics[width=0.25\linewidth]{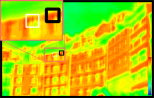}
        \caption{LC}\label{fig:2d}
    \end{subfigure}    
    \begin{subfigure}[t]{\linewidth}
    \centering
    	\includegraphics[width=0.25\linewidth]{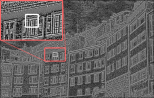}
        \includegraphics[width=0.25\linewidth]{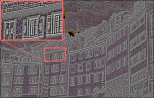}
        \includegraphics[width=0.25\linewidth]{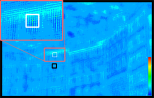}
        \caption{RT}\label{fig:2e}
    \end{subfigure}    
    \begin{subfigure}[t]{\linewidth}
    \centering
    	\includegraphics[width=0.25\linewidth]{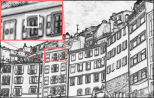}
        \includegraphics[width=0.25\linewidth]{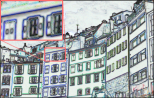}
        \includegraphics[width=0.25\linewidth]{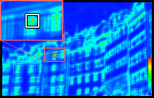}
        \caption{LAT}\label{fig:2f}
    \end{subfigure}    
    \caption[Qualitative results of cross-spectral template matching for Buildings.]{Qualitative results of cross-spectral template matching for Buildings. From left to right transformed NIR image, transformed RGB image, and similarity maps are given. ORG: original NIR and RGB image, HM \cite{HistogramMatching}, LC \cite{ANCC}, RT \cite{Rank}, MTM \cite{MTM} are compared with LAT (ours). White and black rectangular boxes are a sought template and selected window. The figures are best viewed in color.}\label{fig:2}
\end{figure}

\clearpage

The performance of reformulated feature descriptors for cross spectral feature matching is evaluated in this subsection. 100 RGB-NIR image pairs same as previous section were employed for this evaluation. The feature recognition rate is measured for the evaluation and keypoints were detected using SIFT detector. SIFT\cite{SIFT}, BRIEF\cite{BRIEF}, and LSS\cite{LSS} were selected as compared feature descriptors.

Fig. 4.6 shows an example for comparison of LSS and LSS$_{LAT}$. Specifically, in the results, the performance of cross-spectral feature matching are represented with conventional LSS descriptor and LAT-based LSS descriptor, respectively. Note that all the parameters are preserved in all experiments. Since this dataset are structually aligned, reliable correspondence should be also aligned. As shown in the results, LAT-based LSS consistently outperformed the original LSS. Table 4.2 summarizes the recognition rate, showing that the reformulated descriptors show superior performance to the original descriptors. The results show that reformulation with LAT provides promising results for cross spectral feature matching, with an improvement of 10\% recognition rate. 

\begin{figure}[!t]
\renewcommand{\thesubfigure}{}
\centering{
\includegraphics[width=1\linewidth]{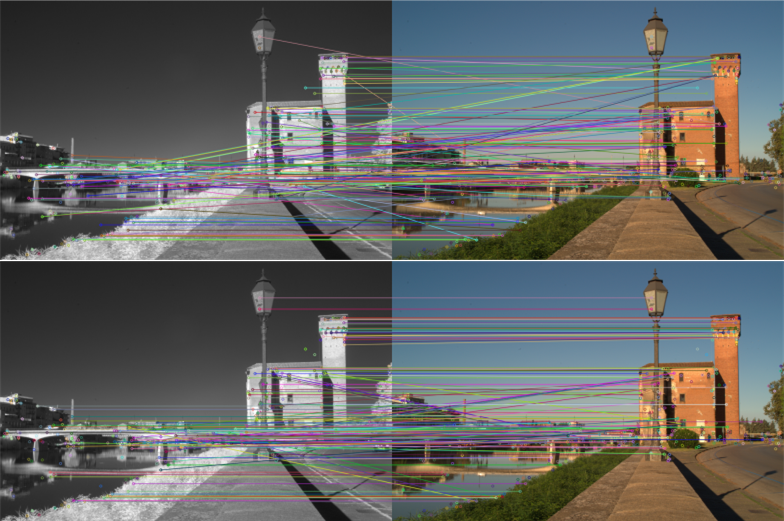}
}
\caption{An example of cross spectral feature matching. Top: LSS and Bottom: LSS$_{LAT}$}
\end{figure}

\begin{table}[!t]
\begin{center}
\caption[Cross-spectral template matching results for RGB-NIR Scene Dataset.]{Cross-spectral template matching results for RGB-NIR Scene Dataset \cite{DB1} in terms of correct detection ratio ($r_{cd}$) and matching pixel error ($e_{mp}$).}
\centering
\setlength{\extrarowheight}{5.0pt}
\begin{tabular}{c|cccccc}
\hline
\hline
Size & ORG & HM & LC & RT & MTM & LAT\\
\hline
$r_{cd}$ & 0.32 & 0.49 & 0.29 & 0.60 & 0.63& \textbf{0.70}\\
$e_{mp}$ & 318 & 218 & 359 & 208 & 158 & \textbf{139}\\
\hline
\end{tabular}
\end{center}
\vspace{1em}
\end{table}

\begin{table}[!b]
\begin{center}
\caption[Cross-spectral feature matching results for RGB-NIR Scene Dataset in terms of recognition rate.]{Cross-spectral feature matching results for RGB-NIR Scene Dataset  \cite{DB1} in terms of recognition rate.}
\centering
\setlength{\extrarowheight}{5.0pt}
\begin{tabular}{c|ccc}
\hline
~ & \multicolumn{3}{c}{Recognition rate}\\
\hline
Original & SIFT \cite{SIFT} & BRIEF \cite{BRIEF} & LSS \cite{LSS}\\
~ & 0.72 & 0.68 & 0.65\\
\hline
LAT & SIFT$_{LAT}$ & BRIEF$_{LAT}$ & LSS$_{LAT}$\\
~ & \textbf{0.85} & \textbf{0.78} & \textbf{0.73}\\
\hline
\end{tabular}
\end{center}
\end{table}

 \subsection{Cross-Radiometry Stereo Matching}


This section provides the superiority of LAT in the task of robust stereo matching in radiometric and photometric deformed stereo images. Stereo matching is commonly formulated as minimization problem of the energy in the MAP-MRF framework \cite{ANCC} as:
\begin{equation}
E(f) = \sum\limits_{\bf{p}} {{D_{\bf{p}}}({f_{\bf{p}}})}  + \sum\limits_{\bf{p}} {\sum\limits_{{\bf{q}} \in \mathcal N_{\bf{p}}} {{V_{{\bf{pq}}}}({f_{\bf{p}}},{f_{\bf{q}}})} },
\end{equation}
where $\mathcal N_{\bf{p}}$ is the neighboring pixels of $\bf{p}$, $f$ is a disparity. In the first term, $D_{\bf{p}}(f_{\bf{p}})$ is the data cost which measures the dissimilarity between $\bf{p}$ in the left image and ${\bf{p}}+f_{\bf{p}}$ in the right image. In the second term, ${{V_{{\bf{pq}}}}({f_{\bf{p}}},{f_{\bf{q}}})}$ is the smoothness cost which penalties non-smooth disparities.

In this experiment, we fixed all of the parameters, cost function, aggregation method, optimization method except for the transformation methods. The absolute difference (AD) for a pixelwise data cost, the adaptive support weight \cite{AdaptiveSupportWieght} with a size of $25\times25$ for the cost aggregation, a truncated quadratic cost for a smoothness cost, and the loopy belief propagation for the global optimization were employed. Although postprocessing like a occlusion-handling and a noise removal can improve the quality of estimated disparities, we did not employ such a postprocessing to more focus on the influence of transform. For the evaluation and comparison of the performance of LAT with others, middlebury stereo data sets \cite{hirschmuller2007evaluation} including Aloe, Baby1, Baby3, Bowling2, Cloth2, Cloth3, Lampshade1, and Monopoly were used. There are three different illumination sources (1,2,3) and three different exposures (indexed as 0,1,2), totally nine different image pairs in each data set. In this experiment, the left image is fixed to illumination source 1 and exposure 1, while the right image is varied in both an illumination and an exposure. In other words, the nine combinations of stereo pairs were used for the evaluation.

Four different methods, HM \cite{HistogramMatching}, LC \cite{ANCC}, RT \cite{Rank}, and CT \cite{census} were employed as compared methods and the original images were also compared as a base method. The qualitative and quantitative comparisons are given in Fig. 4.7 and Table 4.3, respectively. As shown in Table 4.3, the LAT is superior to the other methods in most data sets in terms of bad pixel percentages ($BPP$) and root mean squared errors ($RMSE$). Results presented in Fig. 4.8 and 4.9 show that the qualitative performance of LAT also outperforms the other methods.

\begin{table}[!b]
\begin{center}
\caption[Stereo matching results for illumination and exposure deformed stereo image pairs.]{Stereo matching results for illumination and exposure deformed stereo image pairs in terms of bad pixel percentage ($BPP$) and root mean squared errors ($RMSE$).}
\setlength{\extrarowheight}{5.0pt}
\begin{tabular}{c|cccccc}
\hline
\hline
~ & ORG & HM & LC & RT & CT & LAT\\
\hline
BPP & 0.86 & 0.55 & 0.68  & 0.53 & 0.50 & \textbf{0.47}\\
RMSE & 80.2 & 51.6 & 58.8 & 56.1 & 49.8 & \textbf{45.8}\\
\hline
\end{tabular}
\end{center}
\end{table}

\begin{figure}[!t]
\renewcommand{\thesubfigure}{}
\centering{
\includegraphics[width=0.24\linewidth]{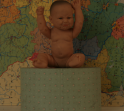}
\includegraphics[width=0.24\linewidth]{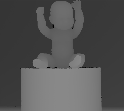}
\includegraphics[width=0.24\linewidth]{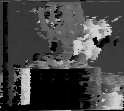}
\includegraphics[width=0.24\linewidth]{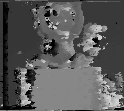}
\includegraphics[width=0.24\linewidth]{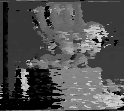}
\includegraphics[width=0.24\linewidth]{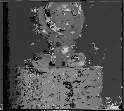}
\includegraphics[width=0.24\linewidth]{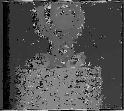}
\includegraphics[width=0.24\linewidth]{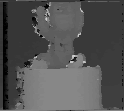}}
\caption[Stereo matching results with a cost function size of 25 for Baby1 stereo pair with Left(1/1) and right(3/1).]{Stereo matching results with a cost function size of 25 for Baby1 stereo pair with Left(1/1) and right(3/1). For each sub-figure, from left-top to right-bottom, figures are left image, ground truth disparity map, disparity map estimated with ORG, HM \cite{HistogramMatching}, LC \cite{ANCC}, RT \cite{Rank}, CT \cite{census}, and LAT, respectively.}
\end{figure}

\begin{figure}[!ht]
\renewcommand{\thesubfigure}{}
\centering
	\begin{subfigure}[t]{\linewidth}
    \centering
    	\includegraphics[width=0.19\linewidth]{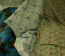}
        \includegraphics[width=0.19\linewidth]{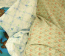}
        \includegraphics[width=0.19\linewidth]{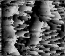}
        \caption{ORG}\label{fig:2a}
    \end{subfigure}
    \begin{subfigure}[t]{\linewidth}
    \centering
    	\includegraphics[width=0.19\linewidth]{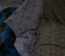}
        \includegraphics[width=0.19\linewidth]{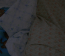}
        \includegraphics[width=0.19\linewidth]{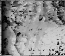}
        \caption{GW}\label{fig:2b}
    \end{subfigure}    
    \begin{subfigure}[t]{\linewidth}
    \centering
    	\includegraphics[width=0.19\linewidth]{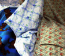}
        \includegraphics[width=0.19\linewidth]{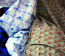}
        \includegraphics[width=0.19\linewidth]{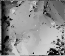}
        \caption{HM}\label{fig:2c}
    \end{subfigure}    
    \begin{subfigure}[t]{\linewidth}
    \centering
    	\includegraphics[width=0.19\linewidth]{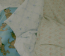}
        \includegraphics[width=0.19\linewidth]{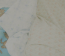}
        \includegraphics[width=0.19\linewidth]{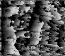}
        \caption{LC}\label{fig:2d}
    \end{subfigure}    
    \begin{subfigure}[t]{\linewidth}
    \centering
    	\includegraphics[width=0.19\linewidth]{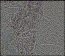}
        \includegraphics[width=0.19\linewidth]{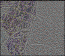}
        \includegraphics[width=0.19\linewidth]{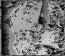}
        \caption{RT}\label{fig:2e}
    \end{subfigure}    
    \begin{subfigure}[t]{\linewidth}
    \centering
    	\includegraphics[width=0.19\linewidth]{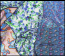}
        \includegraphics[width=0.19\linewidth]{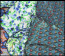}
        \includegraphics[width=0.19\linewidth]{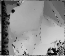}
        \caption{LAT}\label{fig:2f}
    \end{subfigure}    
    \caption[Qualitative results of robust stereo matching for Cloth2.]{Qualitative results of robust stereo matching for Cloth2. Transformed images and disparity map are given.  For each sub-figure, from left to right, figures are left image, right image, disparity map estimated with ORG, HM \cite{HistogramMatching}, LC \cite{ANCC}, RT \cite{Rank}, CT \cite{census}, and LAT, respectively.The figures are best viewed in color.}\label{fig:2}
\end{figure}

\begin{figure}[!ht]
\renewcommand{\thesubfigure}{}
\centering
	\begin{subfigure}[t]{\linewidth}
    \centering
    	\includegraphics[width=0.19\linewidth]{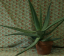}
        \includegraphics[width=0.19\linewidth]{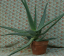}
        \includegraphics[width=0.19\linewidth]{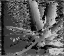}
        \caption{ORG}\label{fig:2a}
    \end{subfigure}
    \begin{subfigure}[t]{\linewidth}
    \centering
    	\includegraphics[width=0.19\linewidth]{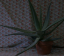}
        \includegraphics[width=0.19\linewidth]{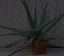}
        \includegraphics[width=0.19\linewidth]{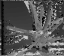}
        \caption{GW}\label{fig:2b}
    \end{subfigure}    
    \begin{subfigure}[t]{\linewidth}
    \centering
    	\includegraphics[width=0.19\linewidth]{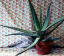}
        \includegraphics[width=0.19\linewidth]{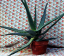}
        \includegraphics[width=0.19\linewidth]{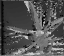}
        \caption{HM}\label{fig:2c}
    \end{subfigure}    
    \begin{subfigure}[t]{\linewidth}
    \centering
    	\includegraphics[width=0.19\linewidth]{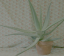}
        \includegraphics[width=0.19\linewidth]{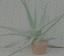}
        \includegraphics[width=0.19\linewidth]{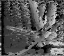}
        \caption{LC}\label{fig:2d}
    \end{subfigure}    
    \begin{subfigure}[t]{\linewidth}
    \centering
    	\includegraphics[width=0.19\linewidth]{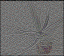}
        \includegraphics[width=0.19\linewidth]{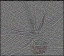}
        \includegraphics[width=0.19\linewidth]{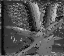}
        \caption{RT}\label{fig:2e}
    \end{subfigure}    
    \begin{subfigure}[t]{\linewidth}
    \centering
    	\includegraphics[width=0.19\linewidth]{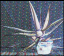}
        \includegraphics[width=0.19\linewidth]{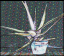}
        \includegraphics[width=0.19\linewidth]{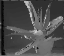}
        \caption{LAT}\label{fig:2f}
    \end{subfigure}    
    \caption[Qualitative results of robust stereo matching for Aloe.]{Qualitative results of robust stereo matching for Aloe. Transformed images and disparity map are given.  For each sub-figure, from left to right, figures are left image, right image, disparity map estimated with ORG, HM \cite{HistogramMatching}, LC \cite{ANCC}, RT \cite{Rank}, CT \cite{census}, and LAT, respectively.The figures are best viewed in color.}\label{fig:2}
\end{figure}

\subsection{Cross-Modality Dense Correspondence Matching}

Estimating visual dense flow from different images but sharing similar scene characteristics is very challenging problem but promising function for a high-level computer vision task \cite{SIFTFLOW}. Especially, cross modality dense flow estimation is more challenging due to their disparate properties \cite{MultiModal}. This section analyzes the performance of SIFT-Flow$_{LAT}$ with a comparison to state-of-the-art methods: SIFT-Flow \cite{SIFTFLOW}, and DAISY \cite{DAISY}\footnote[2]{Since RSNCC \cite{MultiModal} is based on a global matching approach, it is not compared here for fair comparison. SIFT-Flow and DAISY are both based on a local matching approach}. For this purpose, multimodal image database \cite{MultiModal} is employed including RGB-NIR, different exposure, and flash-nonflash image pairs.

Fig. 4.10 shows an qualitative comparison of cross modality dense flow estimated by SIFT-Flow, DAISY, and SIFT-Flow$_{LAT}$. As shown in the figure, compared to the state-of-the-arts methods SIFT-Flow$_{LAT}$ provides a reliable dense flow. Table 4.4 summarizes quantitative comparisons in terms of warping error. The warping error is computed from ground truth displacement for 100 corner points provided in \cite{MultiModal}. The results indicate that SIFT-Flow$_{LAT}$ can be a promising approach for cross modality dense flow estimation.

To address the correspondence-matching problem for different modalities of images, deformation-robust local area transform is proposed. LAT is a nonlinear deformation-invariant transformation of the intensity information into local area information. The experimental results show that LAT and descriptors reformulated by LAT are superior to the conventional methods for matching the correspondence in the context of cross-modality correspondence matching. Specifically, LAT gains approximately a 23\% improvement in correct detection ratio and a 10\% recognition rate increase for the tasks of cross-spectral template matching and feature matching, respectively. LAT also increases the performance of cross-radiation stereo matching and crossmodality dense flow estimation with a 15\% reduction in bad pixel percentage and a 50\% reduction in the warping error, respectively. In conclusion, the local area can be considered as an alternative domain to the intensity domain to achieve robust correspondence matching. Future works should include the development of a cross-modal object recognition based on the properties of LAT

\begin{sidewaysfigure}[!t]
\renewcommand{\thesubfigure}{}
\centering{
\includegraphics[width=0.23\linewidth]{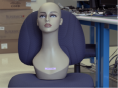}
\includegraphics[width=0.23\linewidth]{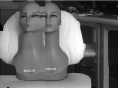}
\includegraphics[width=0.23\linewidth]{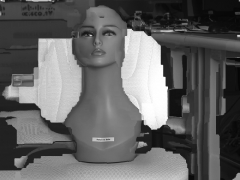}
\includegraphics[width=0.23\linewidth]{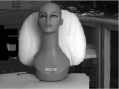}
\includegraphics[width=0.23\linewidth]{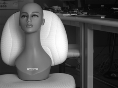}
\includegraphics[width=0.23\linewidth]{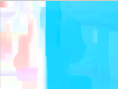}
\includegraphics[width=0.23\linewidth]{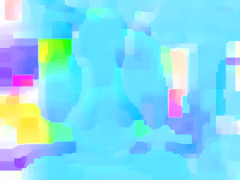}
\includegraphics[width=0.23\linewidth]{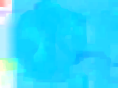}
}
\caption[Cross modality dense flow estimation.]{Cross modality dense flow estimation. From left to right: test image pair, SIFT-flow \cite{SIFTFLOW}, DAISY \cite{DAISY}, SIFT-flow$_{LAT}$. In 2nd, 3rd, 4th column, top images represent warped images and bottom images represent estimated dense flow.}
\end{sidewaysfigure}

\begin{table}[!t]
\begin{center}
\caption[Cross modal dense flow estimation results for multimodal database in terms of warping error.]{Cross modal dense flow estimation results for multimodal database \cite{MultiModal} in terms of warping error.}
\centering
\setlength{\extrarowheight}{5.0pt}
\begin{tabular}{c|cccc}
\hline
Algorithm & RGB-NIR & Flash-Nonflash & Diff. Exp. & All\\
\hline
SIFT-Flow\cite{SIFTFLOW} & 10.11 & \textbf{8.76} & 10.03 & 9.78\\
Variational \cite{VM} & 12.03 & 15.19 & 16.57 & 14.56\\
DAISY \cite{DAISY} & 20.42 & 10.84 & 12.71 & 16.16\\
SIFT-Flow$_{LAT}$ & \textbf{6.83} & 8.83 & \textbf{7.54} & \textbf{7.51}\\
\hline
\end{tabular}
\end{center}
\end{table}


\section{Cross-Modality Deep Scene Recognition}

\ifpdf
    \graphicspath{{Chapter3/Figs/Raster/}{Chapter3/Figs/PDF/}{Chapter3/Figs/}}
\else
    \graphicspath{{Chapter3/Figs/Vector/}{Chapter3/Figs/}}
\fi

\subsection{Cross Spectral Scene Recognition}
In order to study the performance of LAT-Net for cross-spectral scene recognition, we have constructed cross spectral scene database. This database consists of 477 images distributed in 9 categories: Country (52), Field (51), Forest (53), Mountain (55), Old Buildings (51), Street (50), Urban (58), Water (51), where each image is RGB or NIR randomly selected from original RGB-NIR images pairs \cite{DB1}. Randomly selected 99 images were used for testing (11 per category) and remaining 378 images were used for training.  To avoid over-fitting, training images were augmented with resizing (resize ratio is randomly varied from 0.5 - 1.5 with center shift ranged -0.1 - 1.0), rotating (rotating degree is randomly varied from -70 

\clearpage
\noindent - +70 degrees), color-shifted, and flipped. In total, 3,024 training images were employed for training. We trained all networks with the ADAM optimizer \cite{kinga2015method}, learning rate $\eta$=0.001, and batch size $b$=16 for 40 epochs. All networks are pre-trained with places2 database \cite{zhou2017places} for 10 epochs. The places2 is extended version of places dataset \cite{zhou2014learning} and probably the largest scene recognition dataset. In total, the Places2 contains more than 10 million images comprising more than 400 unique scene categories. The dataset includes 5,000 to 30,000 training images per class.

We performed a comparison to state-of-the-art scene recognition methods from hand-crafted methods: GIST \cite{oliva2001modeling}, DiscrimPatches \cite{singh2012unsupervised}, ObjectBank \cite{li2010object} to deep learned feature based methods: fc7-VLAD \cite{gong2014multi}, NetVLAD \cite{arandjelovic2016netvlad}, MFAFVNet \cite{li2017deep}. Table 4.5 presents quantitative comparisons of cross spectral scene recognition in terms of top-1 accuracy. As shown in results, LAT redesigned networks provides highest accuracy even with simple network structure AlexNet \cite{krizhevsky2012imagenet}. LAT-ResNet improved the recognition accuracy by 14.8\% as compared to the state-of-the-arts methods. The results indicate that LAT-redesigned networks is a promising approach for cross spectral scene recognition.

\begin{table}[!t]
\begin{center}
\caption[Cross Spectral Scene Recognition in terms of recognition top-1 accuracy.]{Cross Spectral Scene Recognition in terms of recognition top-1 accuracy. $^1$ hand-crafted feature based methods, $^2$ deep feature based methods, $^3$ holistic deep network based methods}
\centering
\setlength{\extrarowheight}{5.0pt}
\begin{tabular}{c|c}
\hline
Method & Accuracy (\%)\\
\hline
$^1$GIST \cite{oliva2001modeling} & 31.2\\
$^1$1DiscrimPatches \cite{singh2012unsupervised} & 34.2\\
$^1$1ObjectBank \cite{li2010object} & 41.3\\
\hline
$^2$fc7-VLAD \cite{gong2014multi} & 49.4\\
$^2$NetVLAD \cite{arandjelovic2016netvlad} & 53.4\\
$^2$MFAFVNet \cite{li2017deep} & 56.5\\
\hline
$^3$AlexNet \cite{krizhevsky2012imagenet}& 45.4\\
$^3$VGGNet \cite{simonyan2014very} & 51.4\\
$^3$ResNet \cite{he2016deep} & 54.4\\
$^3$LAT-AlexNet (Ours) & \textbf{57.5}\\
$^3$LAT-VGGNet (Ours) & \textbf{65.5}\\
$^3$LAT-ResNet (Ours) & \textbf{69.6}\\
\hline
\end{tabular}
\end{center}
\vspace{1em}
\end{table}



\subsection{Domain Generalized Scene Recognition}
Domain generalization transfers the knowledge learnt from other source domain to an unseen target domain. In order to study the performance of LAT-Net for domain generalized scene recognition, we have conducted the following experiments. All networks are trained on places2 \cite{zhou2014learning} with the ADAM optimizer \cite{kinga2015method}, learning rate $\eta$=0.001, and batch size $b$=16 for 20 epochs. Then, the recognition accuracy is measure on unseen RGB-NIR scene databases. For evaluation, we have constructed three scene databases: RGB, NIR, RGB-NIR combined, which are generated from \cite{DB1}. We divide \cite{DB1} into two separate databases consisting of RGB or NIR, respectively. RGB-NIR combined scene database is same as database employed in Section 4.6.2. Unlike Section 4.6.2, all 477 images are employed as testing images since they are not used for training.

\begin{table}[!t]
\begin{center}
\caption[Domain Generalized Scene Recognition in terms of recognition top-1 accuracy: RGB]{Domain Generalized Scene Recognition in terms of recognition top-1 accuracy: Training on places2 testing on RGB-scene. $^2$ deep feature based methods, $^3$ holistic deep network based methods}
\centering
\setlength{\extrarowheight}{5.0pt}
\begin{tabular}{c|c}
\hline
Method & Accuracy (\%)\\
\hline
$^2$fc7-VLAD \cite{gong2014multi} & 54.3\\
$^2$NetVLAD \cite{arandjelovic2016netvlad} & 58.7\\
$^2$MFAFVNet \cite{li2017deep} & 59.8\\
\hline
$^3$SemanticCluster \cite{george2016semantic} & 66.3\\
\hline
$^3$AlexNet \cite{krizhevsky2012imagenet}& 46.2\\
$^3$VGGNet \cite{simonyan2014very} & 48.3\\
$^3$ResNet \cite{he2016deep} & 51.6\\
$^3$LAT-AlexNet (Ours) & \textbf{58.5}\\
$^3$LAT-VGGNet (Ours) & \textbf{65.4}\\
$^3$LAT-ResNet (Ours) & \textbf{71.2}\\
\hline
\end{tabular}
\end{center}
\vspace{1em}
\end{table}

\begin{table}[!t]
\begin{center}
\caption[Domain Generalized Scene Recognition in terms of recognition top-1 accuracy: NIR]{Domain Generalized Scene Recognition in terms of recognition top-1 accuracy: Training on places2 testing on NIR-scene. $^2$ deep feature based methods, $^3$ holistic deep network based methods}
\centering
\setlength{\extrarowheight}{5.0pt}
\begin{tabular}{c|c}
\hline
Method & Accuracy (\%)\\
\hline
$^2$fc7-VLAD \cite{gong2014multi} & 43.2\\
$^2$NetVLAD \cite{arandjelovic2016netvlad} & 46.8\\
$^2$MFAFVNet \cite{li2017deep} & 49.4\\
\hline
$^3$SemanticCluster \cite{george2016semantic} & 54.7\\
\hline
$^3$AlexNet \cite{krizhevsky2012imagenet}& 36.1\\
$^3$VGGNet \cite{simonyan2014very} & 39.5\\
$^3$ResNet \cite{he2016deep} & 41.6\\
$^3$LAT-AlexNet (Ours) & \textbf{51.9}\\
$^3$LAT-VGGNet (Ours) & \textbf{56.5}\\
$^3$LAT-ResNet (Ours) & \textbf{61.3}\\
\hline
\end{tabular}
\end{center}
\vspace{1em}
\end{table}

We performed a comparison to state-of-the-art scene recognition methods fc7-VLAD \cite{gong2014multi}, NetVLAD \cite{arandjelovic2016netvlad}, MFAFVNet \cite{li2017deep}, and SemanticCluster \cite{george2016semantic}. Table 4.6 and 4.7 present quantitative comparisons of domain generalized scene recognition for RGB and NIR scene databases, respectively, in terms of top-1 accuracy. As shown in results, LAT redesigned networks provides highest accuracy. LAT-ResNet improved the recognition accuracy by 15.9\% and 10.9\% as compared to the state-of-the-arts methods for RGB and NIR scene databases, respectively. The results indicate that LAT-redesigned networks is a promising approach for domain generalized scene recognition. 

\clearpage

%% file: Chapter5/chapter5.tex
\chapter{Conclusion}\label{chap5}

This dissertation proposes deformation-robust image transform, called local area transform (LAT), and mathematically and experimentally prove its invariance properties to nonlinear deformations. LAT is also extended into robust cost functions, feature descriptors, and deep scene recognition networks. 

The experimental results have shown that LAT and descriptors reformulated by LAT were superior to the conventional methods for matching the correspondence in the context of cross-modality correspondence matching. Specifically, LAT gains approximately a 23\% improvement in correct detection ratio and a 10\% recognition rate increase for the tasks of cross-spectral template matching and feature matching, respectively. LAT also increases the performance of cross-radiation stereo matching and cross-modality dense flow estimation with a 15\% reduction in bad pixel percentage and a 50\% reduction in the warping error, respectively. Furthermore, the proposed LAT-Net outperforms existing state-of-the-arts methods in tasks of scene recognition. Specifically, LAT-Net gains up to 14\% accuracy improvement in cross spectral scene recognition task. Also, LAT-Net achieves 6\% and 7\% accuracy improvements for database invariant scene recognition and domain generalized scene recognitions, respectively.

In conclusion, the local area can be considered as an alternative domain to the intensity domain to achieve robust correspondence matching and lots of applications: such as feature matching, stereo matching, dense correspondence matching, and image recognition. we believe the concept of LAT can be extended various potential tasks. Future works include the development of a cross-modal image retrieval and people re-identification based on the properties of local area transformation.